\documentclass[sigconf]{acmart}

\usepackage{balance}
\AtBeginDocument{%
  \providecommand\BibTeX{{%
    \normalfont B\kern-0.5em{\scshape i\kern-0.25em b}\kern-0.8em\TeX}}}

\setcopyright{acmcopyright}\acmConference[MM '22]{Proceedings of the 30th ACM
International Conference on Multimedia}{October 10--14, 2022}{Lisboa, Portugal}
\acmBooktitle{Proceedings of the 30th ACM International Conference on Multimedia
(MM '22), October 10--14, 2022, Lisboa, Portugal}
\copyrightyear{2022}
\acmYear{2022}
\acmPrice{15.00}
\acmDOI{10.1145/3503161.3548338}
\acmISBN{978-1-4503-9203-7/22/10}

\begin{document}

\title{AGTGAN: Unpaired Image Translation for Photographic Ancient Character Generation}

\author{Hongxiang Huang}
\authornote{Both authors contributed equally to this research.}
\author{Daihui Yang}
\authornotemark[1]
\author{Gang Dai}
\authornotemark[1]
\email{eehxhuang@mail.scut.edu.cn}
\affiliation{%
  \institution{South China University of Technology}
  \city{Guangzhou}
  \country{China}
}

\author{Zhen Han}
\affiliation{%
  \institution{Ludwig Maximilian University of Munich}
  \city{Munich}
  \country{Germany}}

\author{Yuyi Wang}
\affiliation{%
  \institution{Swiss Federal Institute of Technology}
  \city{Zurich}
  \country{Switzerland}
  }
\affiliation{%
  \institution{CRRC Institute}
  \city{Zhuzhou}
  \country{China}
  }

\author{Kin-Man Lam}
\affiliation{%
  \institution{The Hong Kong Polytechnic University}
  \city{Hong Kong}
  \country{China}}

\author{Fan Yang}
\affiliation{%
  \institution{South China University of Technology}
  \city{Guangzhou}
  \country{China}}

\author{Shuangping Huang}
\authornote{Corresponding author.}
\affiliation{%
  \institution{South China University of Technology}
  \institution{Pazhou Laboratory}
  \city{Guangzhou}
  \country{China}}

\author{Yongge Liu}
\affiliation{%
  \institution{Anyang Normal University}
  \city{Anyang}
  \country{China}}
  
\author{Mengchao He}
\affiliation{%
  \institution{DAMO Academy, Alibaba Group}
  \city{Hangzhou}
  \country{China}}

\renewcommand{\shortauthors}{Hongxiang Huang et al.}

\begin{abstract}
The study of ancient writings has great value for archaeology and philology. Essential forms of material are photographic characters, but manual photographic character recognition is extremely time-consuming and expertise-dependent. Automatic classification is therefore greatly desired. However, the current performance is limited due to the lack of annotated data. 
Data generation is an inexpensive but useful solution to data scarcity. Nevertheless, the diverse glyph shapes and complex background textures of photographic ancient characters make the generation task difficult, leading to unsatisfactory results of existing methods. To this end, we propose an unsupervised generative adversarial network called AGTGAN in this paper. By explicitly modeling global and local glyph shape style, followed by a stroke-aware texture transfer and an associate adversarial learning mechanism, our method can generate characters with diverse glyphs and realistic textures. 
We evaluate our method on photographic ancient character datasets, e.g., OBC306 and CSDD. Our method outperforms other state-of-the-art methods in terms of various metrics and performs much better in terms of the diversity and authenticity of generated samples. With our generated images, experiments on the largest photographic oracle bone character dataset show that our method can achieve a significant increase in classification accuracy, up to 16.34\%. The source code is available at https://github.com/Hellomystery/AGTGAN.
\end{abstract}

\ccsdesc[500]{Computing methodologies~Image processing}
\ccsdesc[500]{Computing methodologies~Computer vision}

\keywords{ancient character generation, image-to-image translation, GAN}

\maketitle

\begin{figure}
    \begin{center}
       \includegraphics[width=0.9\linewidth]{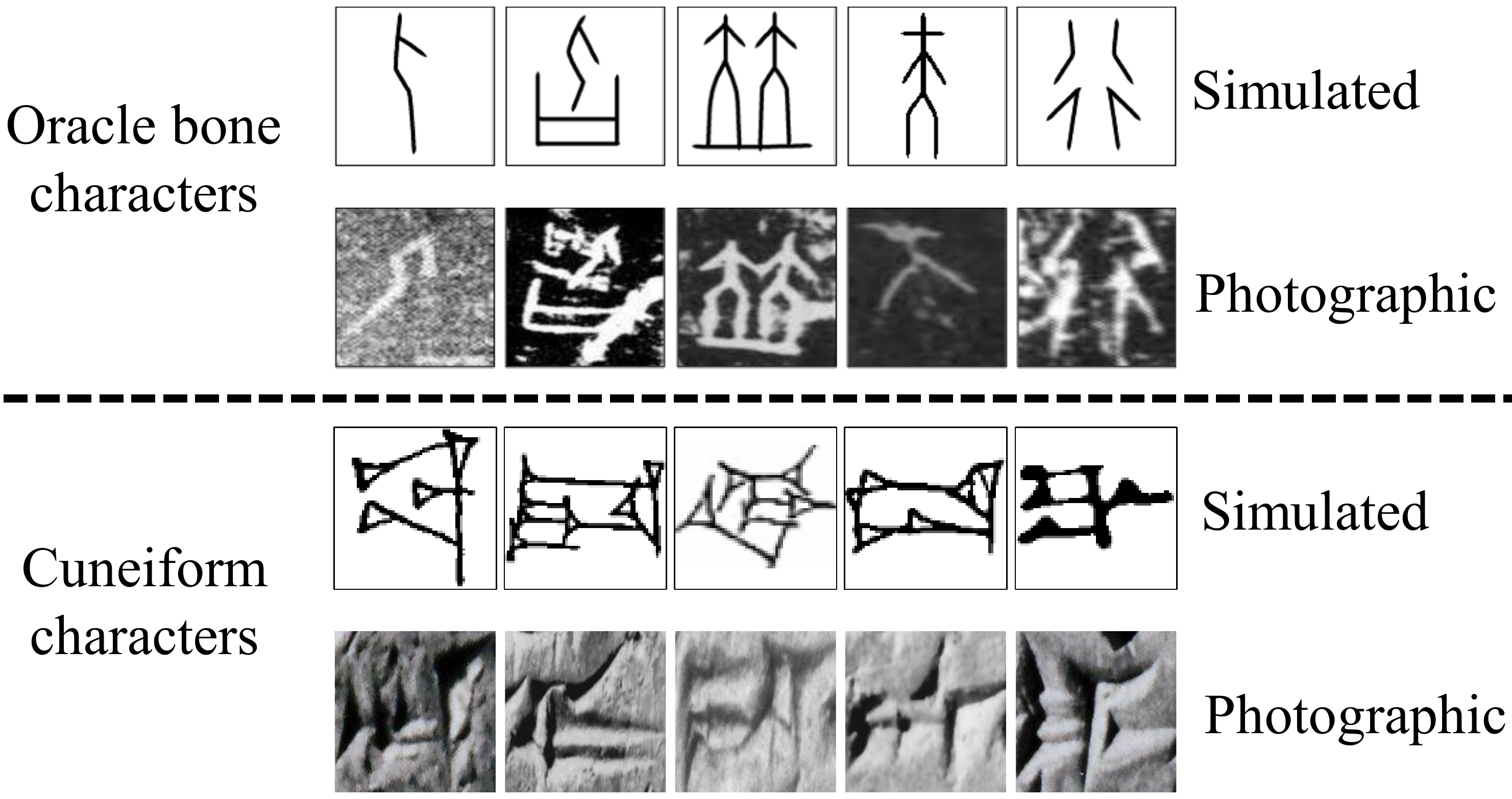}
        \caption{Examples of simulated characters and photographic characters. The first and second rows are simulated and photographic oracle bone characters, respectively. The third and fourth rows are simulated and photographic cuneiform characters, respectively.}
        \label{fig:examples of sc and pc}
    \end{center}
\end{figure}

\section{Introduction}
Human curiosity for exploring the origins of civilization will never fade away even with the passage of time. Through the imprints left by ancestors, i.e., ancient writings, modern people can learn about ancient civilizations. There are several well-known ancient writings, such as oracle bone inscription, cuneiform, Mayan script, Indus script, and ancient Egyptian hieroglyphic. Recognizing the content of these writings is an important premise to boost academic research in both ancient civilizations and literature.

Traditionally, research on ancient writings has been the prerogative of paleographical specialists. Usually, they transform the photographic characters (PC) into simulated characters (SC) by manually simulating the ancient characters on photographic ancient documents. The simulated characters are often included in the ancient character dictionaries~\cite{gates1978outline,budge2012hieroglyphic,Xuzhongshu}, as the primary tools for the study of ancient writings. Fig.~\ref{fig:examples of sc and pc} shows some simulated and photographic characters of oracle bone inscription and cuneiform. However, the specialists identify the photographic characters, based on their experience and intuition, which results in low efficiency and great ambiguity. Automatic photographic ancient character recognition, using deep learning methods, has great potential for accelerating the process of ancient writing research, e.g., decipherment~\cite{hamdany2021translating,panagopoulos2008automatic} and restoration~\cite{sommerschield2020ralegh,zhang2020ai,AssaelSP19}. Nevertheless, deep learning methods require large amounts of well-labeled data to learn an accurate classifier~\cite{zhang2019whole, zhang2020collaborative, xu2020cross, LiuLH21}. Besides, collecting and annotating photographic ancient character data is time-consuming and expertise-dependent. Hence, the performance of existing photographic ancient character classifiers is still unsatisfactory, due to the scarcity of annotated data.

\begin{figure}
    \begin{center}
       \includegraphics[width=0.9\linewidth]{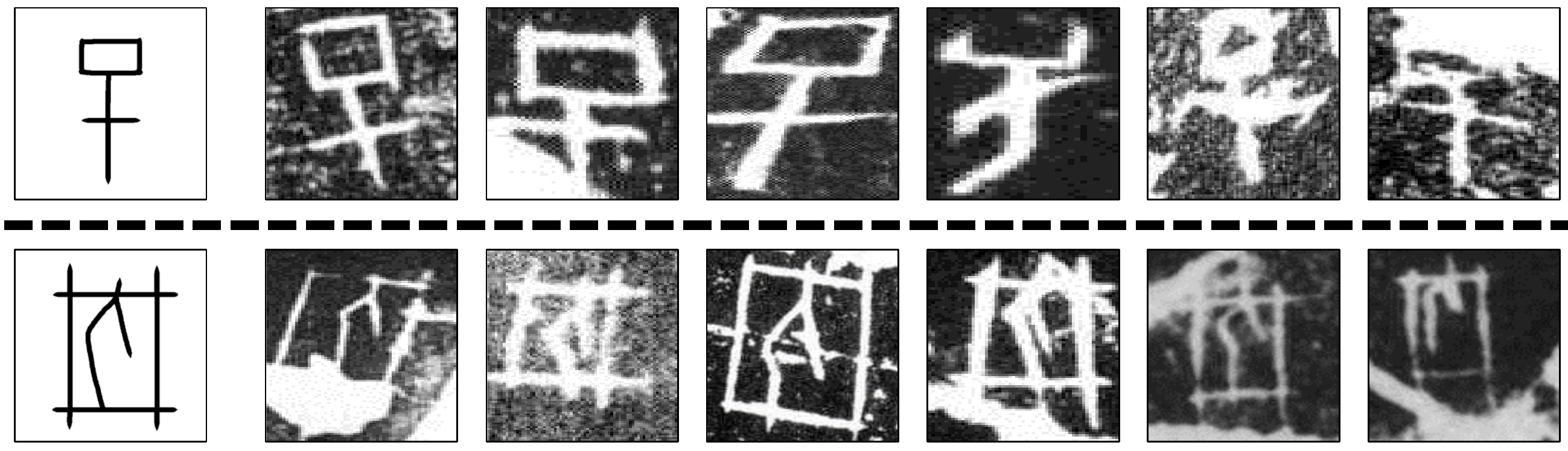}
        \caption{POC examples with diverse glyph shapes and complex background texture.}
        \label{fig:examples of glyph and teture variants}
    \end{center}
\end{figure}

Synthesizing new data is a solution to the data scarcity problem~\cite{guo2019discriminative,suh2022discriminative}. Generative adversarial networks (GANs) \cite{2014Generative} have opened a new door for image generation and brought many amazing results~\cite{ji2019generative, cao2019multi, fogel2020scrabblegan, liu2021deep, liu2021improved}. However, generating photographic ancient characters is not a simple task. Take the photographic oracle bone characters (POC) shown in Fig.~\ref{fig:examples of glyph and teture variants} as an example. Due to the differences in shooting angles and writers, POCs render various global and local shape variations, such as inclination or rotation, changes in the relative position of strokes, etc. In terms of texture, POCs are printed with complex backgrounds, because of irregular noise. Thus, diverse glyph shapes and complex background textures are important aspects to be considered in character generation, which make the task difficult.

The mainstream methods for character generation believe that character images contain content and style information, and styles can be divided into shape style and texture style~\cite{mcgan, gao2019artistic}. Some methods~\cite{xie2021dg, zhang2018separating} learn to extract content and style information in parallel based on disentangled representation, and then entangle the extracted information to generate characters of specific styles. These methods mainly regard the style of a character as its shape style. However, the complex texture of photographic ancient characters increases the difficulty of decoupling the content and style, which leads to degradation of the generation quality when using these methods. In addition, several two-stage methods~\cite{mcgan,jiang2019scfont} divide the character image generation into a shape modeling stage and a stroke rendering stage. Nevertheless, these methods still require a considerable amount of paired data or class labels for training, which is costly and impractical for ancient writings. Moreover, the existing mainstream character generation methods have a common shortcoming that they model shape styles not explicitly enough, which makes it difficult for them to learn diverse global and local shape patterns without strong supervision. To sum up, the generation of photographic ancient characters needs to consider complex shape styles and texture styles. Besides, it is better to reduce the demand for unpaired data and other annotation information through unsupervised learning.

To this end, we propose a novel unsupervised generative model, called Associate Glyph-transformation and Texture-transfer GAN (AGTGAN), which learns a complex mapping from simulated characters to photographic characters, to synthesize diverse and realistic photographic character images. Our model cascades a glyph-transformation GAN (GTG) and a texture-transfer GAN (TTG), and is end-to-end trainable. The contributions of this paper are fourfold:

\begin{enumerate}
    \item The proposed novel photographic character generation model, i.e., AGTGAN, is the first proposed method for enriching annotated photographic character data. This method integrates GTG and TTG, for generating glyph-shape variations and performing texture transfer, respectively. The whole network is end-to-end trainable by the novel associate adversarial training mechanism.
    \item 
    We propose a glyph-shape generator that combines affine with thin-plate-spline (TPS) transformations to explicitly model the global and local shape styles of photographic ancient characters. In addition, noise injection is introduced to increase the randomness of shape transformation, and the signal-and-noise balanced regularization is proposed to guide the model to generate diverse and meaningful glyph shapes.
    \item A new stroke-aware consistency loss for TTG is introduced to solve the blurring problem of the generated photographic characters in the texture transfer process.
    \item Quantitative and qualitative evaluation results show that our generated samples have good diversity and optimal authenticity. 
    With our generated samples, experiments conducted on the largest photographic oracle bone character (POC) dataset, OBC306~\cite{Shuangping2019OBC306}, show that our method achieves an absolute improvement of 16.34\% in terms of POC glyph classification accuracy. 
\end{enumerate}

\section{Related Work}
Character generation has long been considered an essential challenge, while generating photographic ancient characters has not received the attention it deserves. The current generation of ancient writings is limited to a certain type of ancient characters, and the generated samples do not show the original glyph and texture characteristics of the ancient characters~\cite{rusakov2019generating}. 
In this section, we first review the work in similar fields of photographic ancient character generation and then introduce a framework commonly used in character generation.
\subsection{Handwritten Text Generation}
We first review the methods of handwritten text generation, because ancient writings are essentially handwritten or hand-carved texts. The task of handwritten text generation is similar to that of photographic ancient character generation, aiming to imitate natural handwriting in human style. The early two-step methods~\cite{wang2002learning,lin2007style} generate isolated letters, and then concatenate them to produce a whole word. These methods rely on handcrafted rules and only generate handwriting with limited variations. Recently, deep generative models directly generate whole-word images. RNN and GAN were used to generate handwriting in a variety of styles~\cite{graves2013generating,ji2019generative,fogel2020scrabblegan}. 
\cite{kang2020ganwriting} and \cite{bhunia2021handwriting} generate text images on the condition of extracting style features in a few-shot setup and textual content of a predefined fixed length. In addition, \cite{luo2020learn} is an adversarial augmentation method that increases handwritten text images by transforming shapes.
The above-mentioned methods are designed to generate only Latin characters, and most of them require expensive text annotations to enhance the generation quality. 

\subsection{Font generation}

Font generation is usually regarded as an style transfer task for text images, which is handled by image-to-image translation methods in many works. For example, zi2zi~\cite{zi2zi} achieves font style transfer by a conditional GAN. Based on zi2zi, DCFont~\cite{jiang2017dcfont} introduces a style classifier for better style representation. TET-GAN~\cite{yang2019tet} learns to disentangle and recombine the content and style features of text images by a stylization subnetwork and a destylization subnetwork, but it only transfers the texture style. It is difficult to model the complicated shape and texture style at the same time if only relying on a single auto-encoder for modeling the font style. Samaneh et al.~\cite{mcgan} proposed a two-stage font generation framework, MC-GAN, which divides the font style into shape style and texture style and uses two different networks to transform shapes and transfer texture successively. However, the method is only applicable to the 26 English letters, resulting in limited generalization ability. 

For font generation of complex characters, e.g., Chinese characters, the glyph shape transformation has always been the focus of researchers' attention. \cite{zhang2018separating,sun2017learning,wu2020calligan,zeng2021strokegan,huang2020rd,gao2020gan,jiang2019scfont,xie2021dg, wen2021zigan} were proposed for the monochrome Chinese font generation task, which simplify font styles to shape styles. EMD~\cite{zhang2018separating} used different encoders to extract the content vector and shape style vector of fonts based on the idea of disentanglement. SA-VAE~\cite{sun2017learning} demonstrated that domain knowledge of Chinese characters, e.g., the information of radicals and stokes, helps improve the output image quality. CalliGAN~\cite{wu2020calligan} and StokeGAN~\cite{zeng2021strokegan} added extra component codes of characters to train a conditional GAN and a CycleGAN~\cite{zhu2017unpaired}, respectively, exploiting prior knowledge to maintain structural information. RD-GAN~\cite{huang2020rd} proposed a radical extraction module to extract radicals as prior knowledge, which can improve the performance of the discriminator and generate unseen characters in a fixed style. Different from previous methods, ChiroGAN~\cite{gao2020gan} and SCFont~\cite{jiang2019scfont} adopted different font generation paradigms, first extracting the skeleton and then rendering the strokes. However, in DG-Font~\cite{xie2021dg} and ZiGAN~\cite{wen2021zigan}, it was argued that the abovementioned methods~\cite{sun2017learning,wu2020calligan,zeng2021strokegan,huang2020rd,gao2020gan,jiang2019scfont} require expensive supervision information, e.g., stroke information or paired data. DG-Font introduced a feature deformation skip connection, achieved by deformable convolution~\cite{dai2017deformable} to improve the ability of the network to produce shape deformation of the strokes or radicals. ZiGAN learned extra structural knowledge in unpaired data to strengthen the coarse-grained understanding of character content. Inspired by \cite{mcgan} and \cite{zhang2018separating}, AGIS-Net~\cite{gao2019artistic} divided the decoder of the disentanglement framework into two parts, the shape style reconstruction branch and the texture style reconstruction branch, which can realize the shape transformation and texture transfer of complex characters at the same time. 

Nevertheless, existing font generation methods may not be suitable for the generation of photographic ancient characters for the following reasons. On the one hand, the intra-domain glyph shapes of photographic ancient characters are more diverse than those general font generation tasks. The reason is that photographic ancient characters were engraved or written by different people in different periods and photographed from different angles. On the contrary, the shape style of each font is consistent in general font generation tasks, because the characters of each font are written by one writer. On the other hand, although the texture features, e.g., color, brightness and grain, of photographic ancient Chinese characters are relatively stable, it is difficult to decouple the content (glyph) and texture style, because the texture of background noise is very similar to those of foreground strokes, which easily leads to confusion and misunderstanding.


\begin{figure*}
    \begin{center}
        \includegraphics[width=0.8\linewidth]{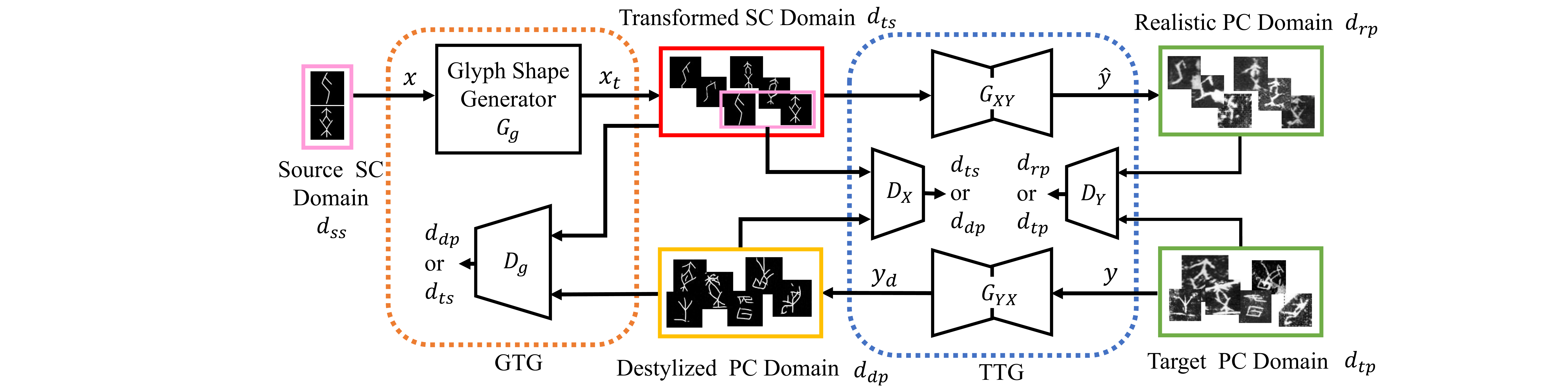}
        \caption{{\bf Method overview.} Given the source SCs, the transformed SC domain is produced by a one-to-many glyph-shape transformation. Then, realistic PCs are generated by texture transfer. In the opposite direction, given the target PCs, a destylized PC domain is produced for guiding  the glyph-shape transformation.}
        \label{fig:method overview}
    \end{center}
\end{figure*}

\subsection{Image-to-Image Translation}
Recently, image-to-image translation has achieved impressive results~\cite{isola2017imagepix2,zhu2017unpaired,lee2020drit++,park2020contrastive,richardson2021encoding}.
\cite{isola2017imagepix2} first proposed a conditional GAN to obtain the desired outputs from reference inputs without manually tuning the mapping function. However, this method requires paired training data. To address the unavailable paired data, \cite{zhu2017unpaired} used a cycle consistency regularization term in the loss function. These two works inspire many follow-up approaches~\cite{park2019semantic,chen2020reusing,han2021dual}. However, these methods only deal with one-to-one translations. To model one-to-many mapping, \cite{lee2020drit++} and \cite{huang2018multimodal} proposed a disentangled representation framework to transfer the source content to a given style. 
In CUT~\cite{park2020contrastive}, it was argued that the cycle-consistency loss used by \cite{zhu2017unpaired,park2019semantic,huang2018multimodal,lee2020drit++,chen2020reusing} assumes that the relationship between the two domains is bijective, which is often too restrictive. An alternative solution was proposed by introducing contrastive learning for unpaired image-to-image translation with a PatchNCE loss to maximize the mutual information between the corresponding patches of the input and output images. Nevertheless, the aforementioned I2I translation methods are capable of texture transfer, but are limited by the shape-variation translation~\cite{improve_Shape_def}. 

To resolve the problem, some methods attempted to model the shape style by directly generating image pixels~\cite{improve_Shape_def, han2021dual}. \cite{improve_Shape_def} proposed a discriminator with dilated convolution~\cite{yu2015multi} to train a shape-aware generator, which achieves global shape transformation. \cite{han2021dual} implemented local geometric transformations using different embedding networks for the comparative learning strategy. \cite{zhan2019spatial} proposed a spatial transformer network (STN)~\cite{jaderberg2015spatial} with thin plate spline (TPS) transformation to explicitly transform scene text images and a CycleGAN to transfer the texture style of the images. However, the discriminator connecting the STN and CycleGAN in~\cite{zhan2019spatial} is unable to discern shape difference under the interference of texture features. These methods only take global or local deformation into account. However, not only global transformation, but also local transformation is needed for generating glyph shapes of ancient writings because the details of glyph shapes, such as strokes and radicals, are diverse.

\section{Proposed Method}
Our goal is to learn a one-to-many mapping from the source simulated character (SC) domain $d_{ss}$ to the target photographic character (PC) domain $d_{tp}$. This mapping should ensure that all multimodal PC outputs preserve the glyph classes of the input SCs, while yielding rich variations in glyph shapes and texture styles.

Instead of establishing a single-step mapping from $d_{ss}$ to $d_{tp}$, we divide this mapping into two stages: glyph shape mapping and texture mapping. This leads to two more intermediate domains, the transformed SC domain $d_{ts}$ and the destylized PC domain $d_{dp}$. Additionally, we define the final generated PC domain as the realistic PC domain $d_{rp}$. All these domains and the overall model are depicted in Fig.~\ref{fig:method overview}, where $x$, $x_{t}$, $y_{d}$, $\hat{y}$, and $y$ denote the characters sampled from $d_{ss}$, $d_{ts}$, $d_{dp}$, $d_{rp}$, and $d_{tp}$, respectively.

As shown in Fig.~\ref{fig:method overview}, our proposed model, called AGTGAN, is composed of a GTG and a TTG for glyph-shape transformation and texture transfer, respectively. Furthermore, we introduce an associate adversarial training mechanism for synergistically improving GTG and TTG, which makes the whole network end-to-end trainable.

\subsection{Glyph-Transformation GAN}
Glyph-transformation GAN (GTG), which consists of a subtly designed glyph-shape generator $G_g$ and a CNN discriminator $D_g$, aims to generate diverse glyph shapes that resemble the PC glyph shapes. 

\subsubsection{Glyph Shape Generator} 
\begin{figure}
\begin{center}
   \includegraphics[width=0.85\linewidth]{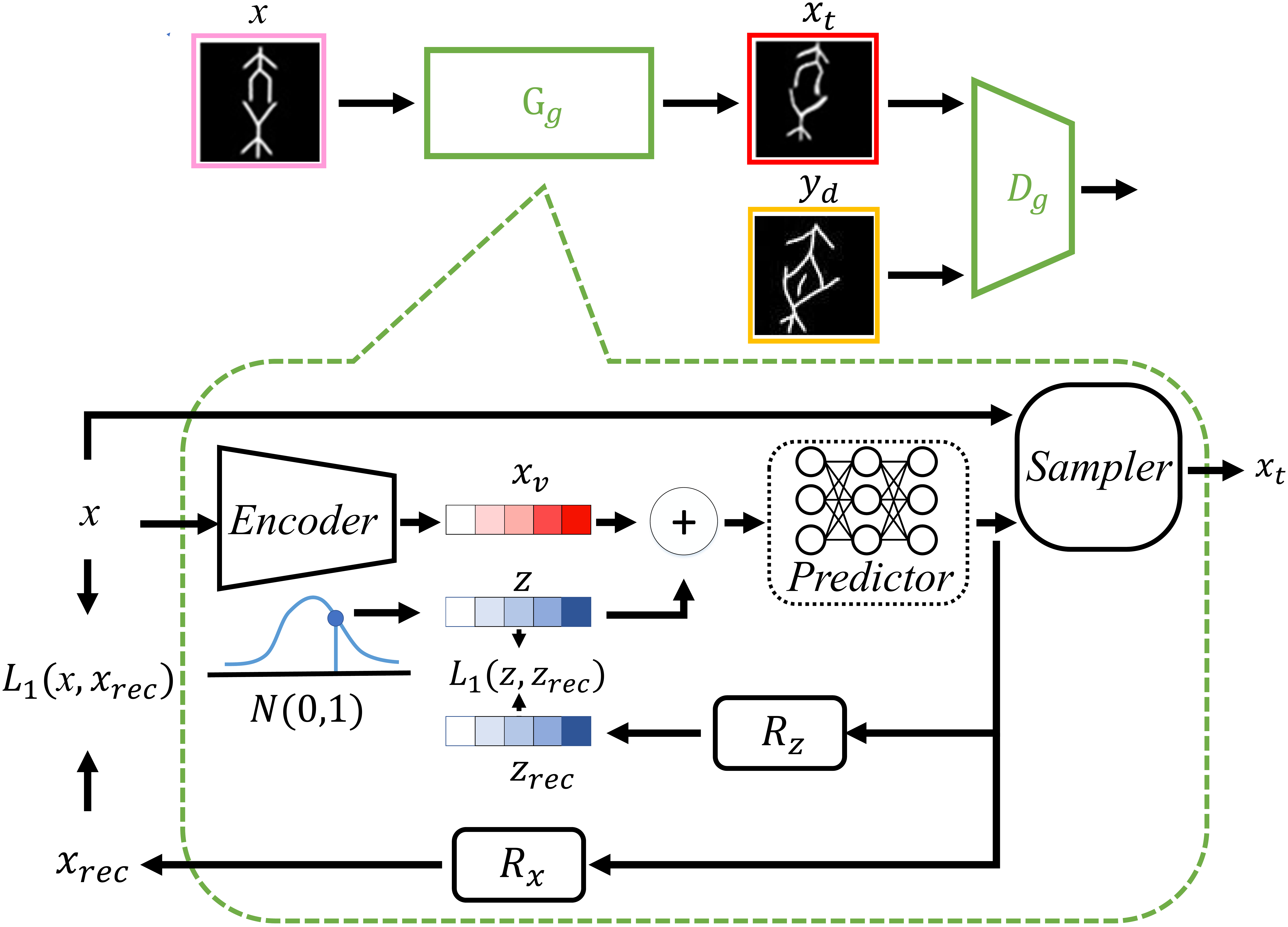}
\end{center}
   \caption{The structure of glyph-transformation GAN.
   }
\label{fig:GTG}
\end{figure}

According to our observations, 
it is difficult to explicitly model glyph shapes by directly predicting
image pixels to achieve global and local shape variations simultaneously without strong supervision. Therefore, we use the spatial transformer network (STN) to resample image pixels with predicted deformed grids, e.g., affine transformation matrix, to achieve explicit shape variations. Our $G_g$, as illustrated in the green dotted box in Fig.~\ref{fig:GTG}, consists of an STN component~\cite{jaderberg2015spatial} and two reconstruction networks, $R_z$ and $R_x$, for reconstructing the noises and the input images, respectively. The STN component 
includes an \textit{Encoder} and a \textit{Predictor}, which are used together to estimate the affine and TPS transformation parameters, and a \textit{Sampler}, which is used to generate $x_{t}$ by resampling the input $x$ with the estimated parameters. Different from those methods that only use affine or TPS transformation~\cite{jaderberg2015spatial,shi2016robust}, our $G_g$ combines affine and TPS transformations, to finely simulate PC glyph shapes at the global and local levels. Specifically, the affine transformation augments the overall glyph shape, producing global shape changes, such as rotation, translation, and scaling, etc., while TPS augments local shape changes, such as stroke length and distortion. We compare the effect of using both TPS and affine transformation with other situations in the Appendix.

To achieve glyph-shape variations, we inject Gaussian noise $z$ to the SC features and then, obtain diversified outputs. 
However, as mentioned in \cite{zhu2017toward}, GAN is prone to ignoring the added noise, thus producing outputs similar to each other. To this end, \cite{zhu2017toward} reconstructed the noise vectors from the outputs, so as to preserve the influence of noise. However, they ignored the fact that noise will reduce the authenticity of the generated images, when the influence of noise far exceeds that of input features. To avoid this, we design two reconstruction networks, $R_z$ and $R_x$, to restore the injected noise $z$ and the input $x$, respectively, from the same estimated parameters during training. With our devised SNR loss (refer to 
the section of `Signal-and-Noise Reconstruction Loss'), $R_{z}$ and $R_{x}$ make the influence of signal and noise compete against each other during training and finally, reach a balanced state.
More details of the architecture of $G_g$ can be found in the Appendix.

\subsubsection{Signal-and-Noise Reconstruction Loss}
As mentioned previously, the signal-and-noise reconstruction (SNR) loss is designed to balance the influence of the input signal and the noise. 
The signal reconstruction loss ensures that the transformed SCs retain the original glyph structure, i.e., the generated SCs and the inputs should belong to the same character classes, while the noise reconstruction loss guarantees the output diversity.

The SNR loss contains three terms. The first two terms correspond to the input signal reconstruction error and the noise reconstruction error,  as follows:
\begin{equation}
    \label{equ:input signal error and noise error terms}
    {\rm L_1}\left(x, x_{rec}\right)+{\rm L_1}\left(z, z_{rec}\right),
\end{equation}%
where $x_{rec}$ and $z_{rec}$ denote the reconstructed signal and noise, respectively, and $L_1$ represents the $L_1$-distance. 
It is worth noting that if the above two terms are not properly regularized, the loss function may be dominant by one of the loss terms. This may lead to either monotonous (not sufficiently) or completely random (meaningless) output.
To address this problem, we introduce a balance term, called reconstruction error ratio ({\rm RER}), to balance the influence of the signal and noise, as follows: 
\begin{equation}
    \label{equ:RER}
    {\rm RER} = \log\left(\frac{{\rm L_1}\left(z, z_{rec}\right)}{{\rm L_1}\left(x,x_{rec}\right)}\right).
\end{equation}%
Including this RER term, the SNR loss is defined as follows:
\begin{small}
    \begin{equation}
        \label{equ:L_r}
        {\rm L_{snr}}\left(G_{g}\right)={\rm L_1}\left(x, x_{rec}\right)+{\rm L_1}\left(z, z_{rec}\right)+\alpha\cdot {\rm RER},
    \end{equation}
\end{small}%
where $\alpha$ is a dynamic coefficient. We further constrain {\rm RER} with a hyperparameter $M>1$. During training, we set $\alpha=1$ if {\rm RER} \textgreater~$\log M$. In this case, the noise reconstruction is much worse than the signal reconstruction. This means that the signal is over-dominant, and we use a positive balance term to penalize a large ratio. On the contrary,  we set $\alpha=-1$ if {\rm RER} \textless~$ -log{M}$. In this case, the signal reconstruction is much worse than the noise reconstruction. This means that the noise is over-dominant, and we use a negative balance term to penalize a small ratio. If {\rm RER} falls between the ideal range, i.e., $[-\log M, \log M]$, we set $\alpha=0$, i.e., without any additional penalty. In this case, the influence of the signal and noise will maintain a balance.

\subsubsection{Diversity Loss}
Diversity is positively correlated with the difference between the two transformation parameters, $P(E(x),z_1)$ and $P(E(x),z_2)$, which are estimated from two mixed signal vectors injected with noises $z_1$ and $z_2$, respectively, where $z_1$ and $z_2$ are randomly drawn from the same Gaussian distribution. The diversity loss is defined as follows:
\begin{small}
    \begin{equation}
        \label{equ:L_div}
        {\rm L_{d i v}}\left(E,P\right)=-{\rm L_1}\left(P\left(E(x), z_{1}\right), P\left(E(x), z_{2}\right)\right),
    \end{equation}
\end{small}%
where $E$ and $P$ represent the \textit{Encoder} and the \textit{Predictor} in STN, respectively.

\subsection{Texture-Transfer GAN}

\label{subsec:Style Transfer GAN}
We use a cycle-structure GAN \cite{zhu2017unpaired}, called texture-transfer GAN (TTG), to add texture styles to SCs, i.e., $x$ and $x_{t}$. TTG consists of two generators, $G_{XY}$ and $G_{YX}$, and two discriminators, $D_Y$ and $D_X$ (see the blue dotted box in Fig.~\ref{fig:method overview}). In order to enhance the adaptability of TTG to PC generation, we propose a stroke-aware cycle consistency loss to prevent PCs from losing strokes or becoming blurred during the texture transfer process.
\subsubsection{Stroke-aware Cycle Consistency Loss}
Compared to other images, the area occupied by the characters in a character image is usually small, and contains almost all the glyph information. This makes existing cycle-structured networks unsuitable for character generation, because they treat every pixel almost equally. Hence, the output is blurred, or even incomplete characters. 
To solve this problem, we propose a stroke-aware cycle consistency loss to guide TTG to pay more attention to foreground characters, as follows:
\begin{small}
    \begin{equation}
        \label{equ:L_cyc}
        \begin{aligned}
            {\rm L_{sacyc}}\left(G_{X Y}, G_{Y X}\right) = \\ &\mathbb{E}_{x_{t}}\left[W \odot {\rm L_{1}}\left(G_{Y X}\left(G_{X Y}\left(\right.\right.\right.\right.\\
            &\left.\left.\left.\left. x_{t}\right)\right),x_{t}\right)\right] \\
            +&\mathbb{E}_{y}\left[{\rm L_1}\left(G_{X Y}\left(G_{Y X}\left(y\right)\right), y\right)\right],
        \end{aligned}
    \end{equation}
\end{small}%
where $\odot$ denotes the element-wise product and $W$ a weight matrix extracted from SC, as follows:
\begin{small}
    \begin{align}
        \label{equ:W_ij}
        W_{ij}=\left\{
            \begin{array}{l}
                \frac{C}{S_{f g}} \cdot S_{b g}, x_{t}^{ij} \in foreground \\
                1,x_{t}^{ij} \in background
            \end{array}\right.,
    \end{align}
\end{small}%
where $S_{fg}$ and $S_{bg}$ denote the area (in pixels) of the foreground region and the background region, respectively. The constant parameter $C\ge 1$ determines the trade-off between enforcing foreground clarity and maintaining background authenticity. It is worth noting that the stroke-aware information does not need additional annotations, which can be easily obtained by image binarization.

\subsection{Associate Adversarial Training}

We propose an associate adversarial training, which connects GTG and TTG through two transitional domains, $d_{ts}$ and $d_{dp}$, to exchange the glyph-shape information and fuse it with texture styles, as shown in Fig.~\ref{fig:method overview}. In this way, the two GANs are trained jointly to adapt to and promote each other. 

To implement the glyph-shape mapping from the source SC domain $d_{ss}$ to the target PC domain $d_{tp}$, $G_g$ of the first GAN attempts to explore diverse glyph shapes, while $D_g$ attempts to distinguish whether the glyph shapes come from the transformed SC domain, ${d_{ts}}$, or the destylized PC domain, $d_{dp}$. $d_{dp}$ is produced via the second GAN with a cycle structure, acting as a bridging domain between the two GANs and guiding the glyph deformation of the input SCs. Through adversarial learning, GTG pushes the distribution of $d_{ts}$ close to that of $d_{dp}$.

We use the transitional domain $d_{dp}$, instead of the target PC domain $d_{tp}$, to guide the glyph transformation, because there are some texture style differences between $d_{tp}$ and $d_{ss}$. This will confuse GTG in the glyph transformation learning. $d_{dp}$ is obtained after gradual destylization through training the TTG, retaining rich global and local glyph-shape patterns. When the texture style of $d_{dp}$ is gradually reduced, GTG focuses on the glyph-shape differences and precisely learns the glyph shape features.

The loss function for GTG is a combination of the least squares generative adversarial loss~\cite{LSGAN},  the SNR loss, and the diversity loss, as follows:
\begin{small}
    \begin{align}
        \label{equ:L_Gg}
        \begin{split}
            {\rm L_{G_g}}=&\mathbb{E}_{x }\left[D_g\left(G_{g}{(x)}\right)-1\right]^{2}\\
            &+ {\rm L_{snr}}\left(G_{g}\right)+{\rm L_{div}}\left(E,P\right),
        \end{split}& \\
        \label{equ:L_Dg}
        \begin{split}
            {\rm L_{D_g}}=&\frac{1}{2} \mathbb{E}_{x}\left[D_g\left(G_g\left(x\right)\right)\right]^{2}\\
            &+\frac{1}{2}\mathbb{E}_{y_{d}}\left[D_g\left(y_{d}\right)-1\right]^{2}.
        \end{split}&
    \end{align}
\end{small}%
To perform the style mapping from the source SC domain $d_{ss}$ to the target PC domain $d_{tp}$, $G_{XY}$ takes SCs from $d_{ts}$ as its input and generates realistic PCs to deceive the corresponding $D_Y$. $D_Y$ attempts to distinguish the domain of the PC samples, i.e., either the realistic PC domain $d_{rp}$ or the target PC domain $d_{tp}$.

\begin{figure*}[t]
\begin{center}
    \includegraphics[width=0.8\linewidth]{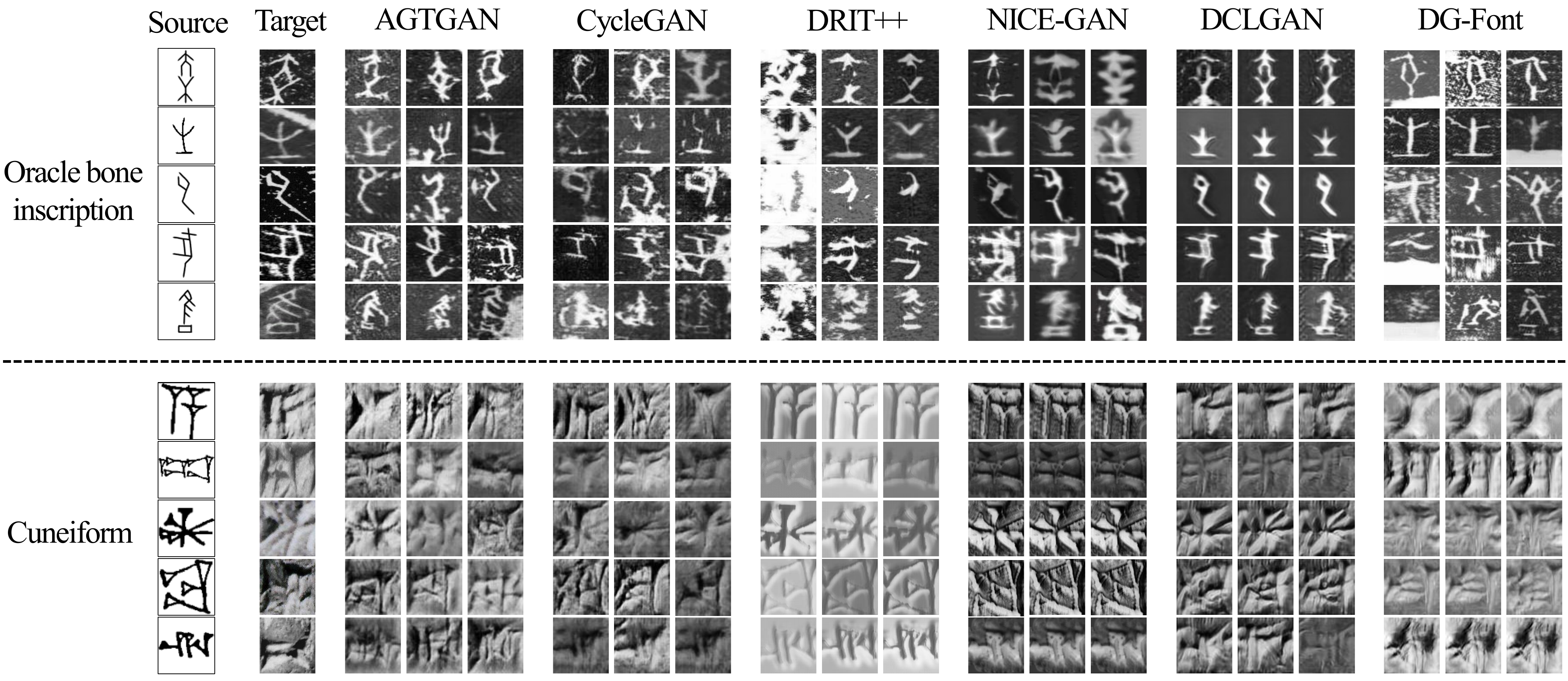}
\end{center}
   \caption{Generated POC and PCC images. The first column is the source SC images, the second column is the target PC images, and the other columns are generated by different methods.}
\label{fig:generated PC image}
\end{figure*}

We use $d_{ts}$, instead of $d_{ss}$, as the input domain for TTG, because the glyph difference between $d_{ss}$ and $d_{tp}$ can easily confuse TTG when learning to transfer texture styles. With GTG training, the glyph shapes of the $d_{ts}$ samples gradually approach the glyph of $d_{dp}$. In this way, TTG can focus on the texture style differences and capture texture features more accurately. GTG transmits glyph-shape information to TTG by $d_{ts}$, which encourages $G_{XY}$ to generate PCs with realistic glyph shapes and texture styles.

The loss function for TTG is a combination of the least squares generative adversarial loss and the stroke-aware cycle consistency loss, as follows:
\begin{small}
    \begin{align}
        \label{equ:L_GXY_GYX}
        \begin{split}
            {\rm L_{G_{XY},G_{YX}}}&=\mathbb{E}_{x_{t}}\left[D_Y\left(G_{XY}{(x_{t})}\right)-1\right]^{2}\\
                &+\mathbb{E}_{y}\left[D_{X}\left(G_{YX}{(y)}\right)-1\right]^{2}\\
                &+\lambda {\rm L_{sacyc}}\left(G_{XY} , G_{Y X}\right),
        \end{split}& \\
        \label{equ:L_DY_D_X}
        \begin{split}
            {\rm L_{D_{Y},D_{X}}}=&\frac{1}{2}\mathbb{E}_{x_{t}}\left[D_Y\left(G_{XY}\left(x_{t}\right)\right)\right]^{2}\\
            &+\frac{1}{2}\mathbb{E}_{y}\left[D_Y\left(y\right)-1\right]^{2}\\
            &+\frac{1}{2} \mathbb{E}_{y}\left[D_X\left(G_{YX}\left(y\right)\right)\right]^{2}\\
            &+\frac{1}{2}\mathbb{E}_{x}\left[D_X\left(x\right)-1\right]^{2},
        \end{split}&
    \end{align}
\end{small}%
where $\lambda$ is a hyperparameter, which controls the relative importance of the stroke-aware cycle consistency loss.

With the proposed associate adversarial training for GTG and TTG, they can be trained together harmoniously as a unified pipeline, i.e., AGTGAN.

\section{Experiment}
\subsection{Datasets}

\noindent\textbf{SOC5519}~\cite{soc_from} contains 44,868 clean simulated oracle bone character (SOC) instances from 5,491 classes, covering almost all the classes that have been discovered. 

\noindent\textbf{OBC306}~\cite{Shuangping2019OBC306} contains 309,511 samples from 306 classes, available from the open-source OBI database.
It is worth noting that most of the character classes in SOC5519 are unavailable in OBC306 since the number of classes in OBC306 is much fewer than that in SOC5519. 

\noindent\textbf{HCCC}~\cite{Yamauchi:2018} is an image set that was manually simulated and arranged from existing hand copies of cuneiform tablets. This dataset contains 4,416 samples from 50 classes that most frequently appear in several corpora.

\noindent\textbf{CSDD}~\cite{dencker2020deep} provides bounding box annotations and class labels for signs on 81 tablets' images. We segmented 2,576 photographic cuneiform character (PCC) images according to the bounding box annotations over 233 classes.

\subsection{Training Details}
For all experiments, we set $\lambda = 10$, $C=2$, and $M = 6$. All the simulated character and photographic images are resized to $64\times64$. We set the initial learning rate at 0.0001 for GTG and 0.001 for TTG, and use the Adam solver~\cite{kingma2014adam} with a batch size of 64 for optimization. We keep the learning rates constant for the first 15,000 iterations, and linearly decay the rates to zero over the next 15,000 iterations.

\subsection{Quantitative Evaluation Metrics \& User Study}

\noindent\textbf{FID.} 
we use the Fréchet inception distance (FID)~\cite{heusel2017gans} to evaluate the authenticity of the generated images by measuring the distance between the generated distribution and the real distribution based on the features extracted by the inception network~\cite{incepv42017}. The lower the FID, the better the quality of the generated images.

\noindent\textbf{NDB and JSD.}
We use the number of statistically different bins (NDB) and the Jensen-Shannon divergence (JSD)~\cite{richardson2018gans} to evaluate the authenticity of the generated images. Compared with FID, the NDB and JSD metrics are directly applied to image pixels and do not rely on the learned representation. This makes the metric more sensitive to pixel-level differences in images. We set the number of bins for NDB to 50.

\noindent\textbf{LPIPS.} 
We use the learned perceptual image patch similarity (LPIPS) metric~\cite{zhang2018unreasonable} to measure the diversity of generated images. We generate 1,000 samples for each class and compute the LPIPS distance between pairs of samples. The average LPIPS distance for all the classes is used as the final LPIPS value.

\noindent\textbf{User Study.}
We conduct a user study based on pairwise comparisons. Given the POC image groups generated by AGTGAN and other models, each subject needs to answer the question ``Which POC image group is more realistic?'' with a real POC image as a reference.

\subsection{Comparison with State-of-the-Art Methods}
\label{subsection:Comparison with State-of-the-Arts}
We compare out proposed AGTGAN with state-of-the-art unsupervised image-to-image models, including CycleGAN~\cite{zhu2017unpaired}, DRIT++~\cite{lee2020drit++}, NICE-GAN~\cite{chen2020reusing}, DCLGAN~\cite{han2021dual}, and DG-Font~\cite{xie2021dg}, from the perspectives of visual quality, quantitative metrics, user study, and classification performance. 

\subsubsection{Generation Result}

\begin{table}
    \caption{Quantitative Evaluation of Generated POCs.}
    \begin{center}
            \begin{tabular}{ccccc} 
                \hline
                Methods     & FID$\downarrow$ & NDB$\downarrow$  & JSD$\downarrow$ & LPIPS$\uparrow$ \\ 
                \hline
                DCLGAN      & 176.46 & 41 & 0.178 & 0.234   \\
                DG-Font     & 168.93 &  39  & 0.123 & 0.224       \\
                NICE-GAN      & 140.22 & 41 & 0.104 &  {\bf 0.325}\\
                CycleGAN      & 130.76 & 35 & 0.135 & 0.252\\
                DRIT++        & 117.03 & 37 & 0.083 & 0.299\\ 
                AGTGAN       & {\bf 99.48} & {\bf 26} & {\bf 0.066}  & 0.284\\ 
                \hline
            \end{tabular}
    \end{center}
    \label{tab:quantitative evaluation}
\end{table}

\begin{figure}
    \centering
    \includegraphics[width=0.7\linewidth]{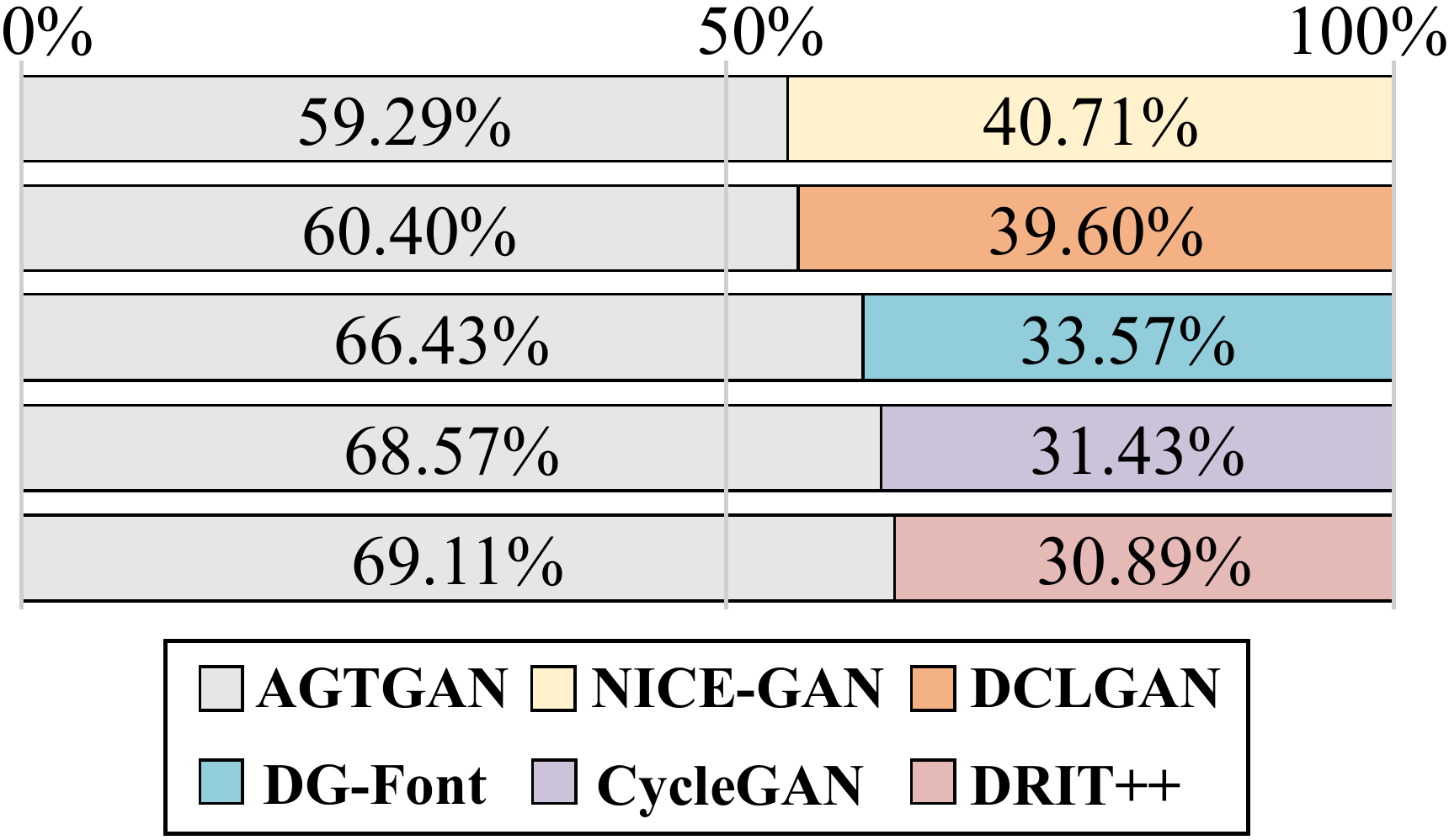}
    \caption{The authenticity of the generated POCs by human evaluation.
    The numbers indicate the percentage of preference for the comparison pairs.}
    \label{fig:human evaluation result}
\end{figure}

\begin{table}
    \caption{The average class accuracy achieved by different methods. ``Source only'' refers to training the recognizer without using generated incremental images.}
    \begin{center}
            \begin{tabular}{cccc} 
        \hline
                 & \multicolumn{3}{c}{TOP-1(\%)} \\ 
                              \cline{2-4} 
        Methods  & OBC306  & FS  & ZS \\ 
                               \hline
        Source only   & 69.02       & 5.36    & 0 \\
        DG-Font       & 75.11       & 38.87        & 27.45   \\
        CycleGAN         & 78.07      & 54.46        & 48.28 \\
        DCLGAN              &80.77      &58.33           &54.90 \\
        DRIT++           & 81.04     & 57.14        & 51.72 \\
        NICE-GAN          & 83.62       &68.75           &79.31 \\
        AGTGAN(Ours)     & {\bf 85.36}        & {\bf 81.25}     & {\bf 93.10} \\ 
                                \hline
       \end{tabular}
    \end{center}
    \label{tab:class average accuracy}
\end{table}

Fig.~\ref{fig:generated PC image} demonstrates some generated photographic ancient characters for randomly selected character classes. 
We can see that our method generates significantly diverse glyph shapes: the strokes and radicals show rich local variations, and the entire characters show different sizes or inclination appearances, while preserving the original glyph labels. In terms of texture style, our generated POC renders more natural background noise and the generated PCC renders a better three-dimensional effect. Furthermore, the characters of both oracle bone inscription and cuneiform generated by our method do not suffer from blurring.

In contrast, CycleGAN generates some implausible glyphs, e.g., some strokes are missing or wrong strokes are added. This is mainly because the transferred background texture confuses the character strokes, leading to blurring of the generated glyphs. DG-Font, DRIT++, and NICE-GAN generate worse results, such as unrecognizable glyphs and images with artifacts or fog effect. 
A possible reason is that they rely on the decoupled latent vectors of content and style, which are not easily obtained from photographic characters with complex glyph shapes and textures.
Although DCLGAN generates clear glyphs, the background texture of the generated samples is very monotonous and is quite different from the target texture styles. The reason is that the adopted contrastive learning strategy is prone to capturing semantic information of glyphs, while ignoring most of the background details.
In addition, many generated samples, e.g., in the second and third rows under ``DRIT++'', ``NICE-GAN'' and ``DCLGAN'', still align with the source characters with less glyph-shape variations, indicating that the baselines have difficulties in learning the variety of glyphs from the target. Although DG-Font uses deformable convolution to learn shape styles, it can only produce weak local deformation. It is worth mentioning that we demonstrate the great potential of our method in zero-shot generation in the Appendix.

\begin{figure*}
\begin{center}
   \includegraphics[width=0.8\linewidth]{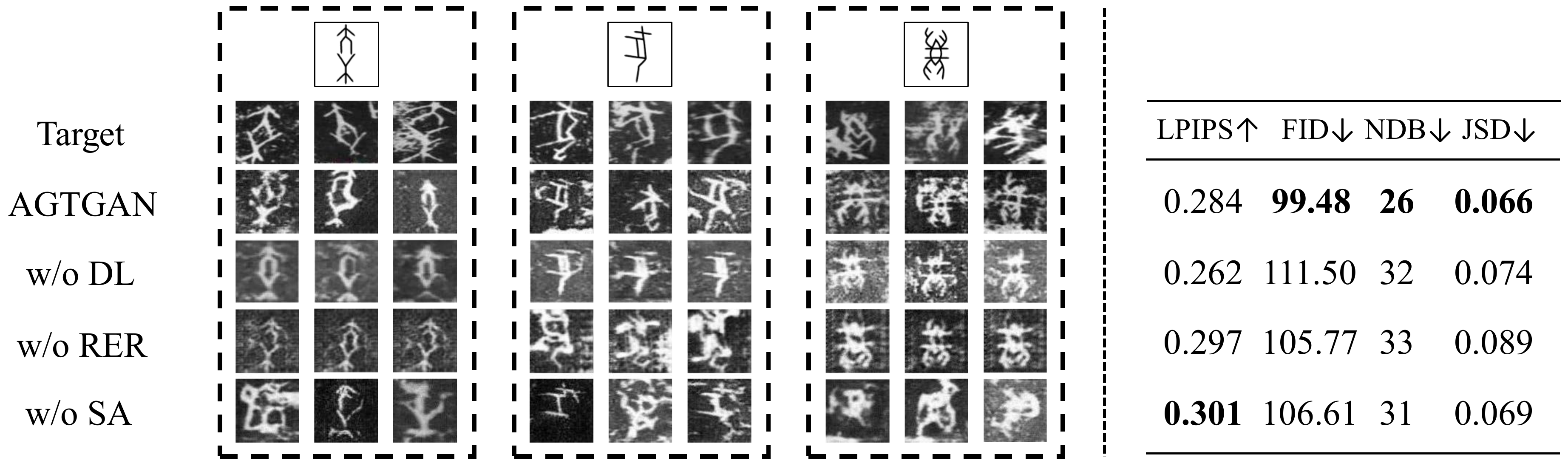}
\end{center}
   \caption{Ablation study of different parts of AGTGAN. We show the source SOCs (1st row), target POCs (2nd row), generated samples of AGTGAN (3rd row), and generated samples of AGTGAN without DL (4th row), without RER (5th row), and without SA (6th row). 
   }
\label{fig:ablation study}
\end{figure*}

Tab.~\ref{tab:quantitative evaluation} summarizes the quantitative results, and similar conclusions to the above visual analysis can be reached. Our method achieves the best FID, NDB and JSD scores among all methods. 
Although DRIT++ and NICE-GAN obtain a higher LPIPS score than our method, the high diversity comes from chaotic textures and incomplete glyphs.  It is worth noting that the classification criteria of the HCCC and CSDD datasets are inconsistent and the samples generated by reference to HCCC cannot be evaluated with the real data of CSDD when applying the FID metric. The results of LPIPS, NDB, and JSD applied to the generated cuneiform also show that our method is superior to other comparison methods, which can be found in the Appendix.

The results of the user study in Fig.~\ref{fig:human evaluation result} show that more people choose the POCs generated by our AGTGAN to be closer to the target POCs, compared with other methods.

\subsubsection{Classification Performance}

We further evaluate the quality of the generated samples by conducting classification experiments on OBC306. Following the same protocol as in~\cite{Shuangping2019OBC306}, we randomly select a quarter of the samples for testing while ensuring that each class has at least one test sample, and the rest form the real POC training set. 
More setting details can be found in the Appendix.
We also measured the classification performance of the minority classes to further investigate the effect of the generated POCs.
Those classes that contain one to ten samples form a few-shot (FS) subset, and those that do not contain any real training samples form a zero-shot (ZS) subset.

Tab.~\ref{tab:class average accuracy} summarizes the TOP-1 average class accuracy. The TOP-3 and TOP-5 accuracies are listed in the Appendix. Tab.~\ref{tab:class average accuracy} shows that, with the AGTGAN generated samples, the highest POC classification accuracy can be achieved on OBC306, as well as FS and ZS. 
Compared to ``Source only'' in Tab.~\ref{tab:class average accuracy}, the improvement achieved using our generated samples is at least 16.34\%. By using the training data generated by AGTGAN, the classification accuracy is 81.25\% on FS and 93.10\% on ZS. The results show that our method has a significant effect on improving the recognition accuracy of the minority classes.
 
\subsection{Ablation Study}
Fig.~\ref{fig:ablation study} summarized the ablation studies for evaluating the impact of different parts of AGTGAN. 

{\bf Diversity loss (w/o DL)}:
Our model, without using the diversity loss, produces POC images with monotonous glyph patterns, which are almost aligned with the SOC glyphs. Thus, the diversity of the generated POCs is not satisfactory. This can also be seen from its smallest LPIPS value, i.e., 0.262, among all the methods compared.

{\bf Reconstruction error ratio (w/o RER)}:
From the row of ``w/o RER'' in Fig.~\ref{fig:ablation study}, we can see two distinct types of generated glyphs. One is with little transformation, e.g., the first and third glyph classes, and the other is with large distortion, beyond the plausible glyph space, e.g., the second glyph class. This is because the balance between signal and noise may be broken in the optimization process, if {\rm RER} is not imposed for balancing. This may cause the noise to become small, which is then filtered by the network, or to become large and dominate the signal.

{\bf Stroke-aware cycle consistency loss (w/o SA)}: 
The “w/o SA” row in Fig.~\ref{fig:ablation study} shows that the generated samples have missing strokes or the background textures are incorrectly added to the foreground characters. This dramatically degrades the generation quality. The LPIPS of the generated samples reaches a high value of 0.301, which is mainly due to chaotic glyphs rather than plausible variations.

\section{Conclusion}
We proposed a novel character generative model, namely AGTGAN, which, so far as we know, is the first method capable of automatically generating rich and realistic photographic ancient characters. 
Hence, these generated images, to a certain extent, solve the most critical problem faced by the task of automatic classification of photographic ancient characters, due to lack of well-labeled data. 


A natural direction for future work is to extend our proposed method to more general and complex writing, e.g., handwritten text and formulas. Moreover, the creative font generation tasks, e.g., font design and calligraphy imitation, are other potential applications of our proposed method. 

\section{Acknowledgements}
The research is partially supported by National Nature Science Foundation of China (No. 62176093, 61673182, 61936003), Key Realm R\&D Program of Guangzhou (No. 202206030001), Guangdong Basic and Applied Basic Research Foundation (No. 2021A1515012282), GD-NSF (No. 2017A030312006), and the Alibaba Innovative Research.

\bibliographystyle{ACM-Reference-Format}
\balance
\bibliography{egbib}


\begin{thebibliography}{67}


\ifx \showCODEN    \undefined \def \showCODEN     #1{\unskip}     \fi
\ifx \showDOI      \undefined \def \showDOI       #1{#1}\fi
\ifx \showISBNx    \undefined \def \showISBNx     #1{\unskip}     \fi
\ifx \showISBNxiii \undefined \def \showISBNxiii  #1{\unskip}     \fi
\ifx \showISSN     \undefined \def \showISSN      #1{\unskip}     \fi
\ifx \showLCCN     \undefined \def \showLCCN      #1{\unskip}     \fi
\ifx \shownote     \undefined \def \shownote      #1{#1}          \fi
\ifx \showarticletitle \undefined \def \showarticletitle #1{#1}   \fi
\ifx \showURL      \undefined \def \showURL       {\relax}        \fi
\providecommand\bibfield[2]{#2}
\providecommand\bibinfo[2]{#2}
\providecommand\natexlab[1]{#1}
\providecommand\showeprint[2][]{arXiv:#2}

\bibitem[Assael et~al\mbox{.}(2019)]%
        {AssaelSP19}
\bibfield{author}{\bibinfo{person}{Yannis~M. Assael}, \bibinfo{person}{Thea
  Sommerschield}, {and} \bibinfo{person}{Jonathan Prag}.}
  \bibinfo{year}{2019}\natexlab{}.
\newblock \showarticletitle{Restoring ancient text using deep learning: a case
  study on Greek epigraphy}. In \bibinfo{booktitle}{\emph{EMNLP-IJCNLP}}.
  \bibinfo{pages}{6367--6374}.
\newblock


\bibitem[Azadi et~al\mbox{.}(2018)]%
        {mcgan}
\bibfield{author}{\bibinfo{person}{Samaneh Azadi}, \bibinfo{person}{Matthew
  Fisher}, \bibinfo{person}{Vladimir~G Kim}, \bibinfo{person}{Zhaowen Wang},
  \bibinfo{person}{Eli Shechtman}, {and} \bibinfo{person}{Trevor Darrell}.}
  \bibinfo{year}{2018}\natexlab{}.
\newblock \showarticletitle{Multi-content gan for few-shot font style
  transfer}. In \bibinfo{booktitle}{\emph{CVPR}}. \bibinfo{pages}{7564--7573}.
\newblock


\bibitem[Bhunia et~al\mbox{.}(2021)]%
        {bhunia2021handwriting}
\bibfield{author}{\bibinfo{person}{Ankan~Kumar Bhunia}, \bibinfo{person}{Salman
  Khan}, \bibinfo{person}{Hisham Cholakkal}, \bibinfo{person}{Rao~Muhammad
  Anwer}, \bibinfo{person}{Fahad~Shahbaz Khan}, {and} \bibinfo{person}{Mubarak
  Shah}.} \bibinfo{year}{2021}\natexlab{}.
\newblock \showarticletitle{Handwriting transformers}. In
  \bibinfo{booktitle}{\emph{Proceedings of the IEEE/CVF International
  Conference on Computer Vision}}. \bibinfo{pages}{1086--1094}.
\newblock


\bibitem[Budge(2012)]%
        {budge2012hieroglyphic}
\bibfield{author}{\bibinfo{person}{EA~Wallis Budge}.}
  \bibinfo{year}{2012}\natexlab{}.
\newblock \bibinfo{booktitle}{\emph{Hieroglyphic Vocabulary to the Book of the
  Dead}}.
\newblock \bibinfo{publisher}{Courier Corporation}.
\newblock


\bibitem[Cao et~al\mbox{.}(2019)]%
        {cao2019multi}
\bibfield{author}{\bibinfo{person}{Jiezhang Cao}, \bibinfo{person}{Langyuan
  Mo}, \bibinfo{person}{Yifan Zhang}, \bibinfo{person}{Kui Jia},
  \bibinfo{person}{Chunhua Shen}, {and} \bibinfo{person}{Mingkui Tan}.}
  \bibinfo{year}{2019}\natexlab{}.
\newblock \showarticletitle{Multi-marginal wasserstein gan}.
\newblock \bibinfo{journal}{\emph{Advances in Neural Information Processing
  Systems}}  \bibinfo{volume}{32} (\bibinfo{year}{2019}).
\newblock


\bibitem[Dai et~al\mbox{.}(2017)]%
        {dai2017deformable}
\bibfield{author}{\bibinfo{person}{Jifeng Dai}, \bibinfo{person}{Haozhi Qi},
  \bibinfo{person}{Yuwen Xiong}, \bibinfo{person}{Yi Li},
  \bibinfo{person}{Guodong Zhang}, \bibinfo{person}{Han Hu}, {and}
  \bibinfo{person}{Yichen Wei}.} \bibinfo{year}{2017}\natexlab{}.
\newblock \showarticletitle{Deformable convolutional networks}. In
  \bibinfo{booktitle}{\emph{Proceedings of the IEEE international conference on
  computer vision}}. \bibinfo{pages}{764--773}.
\newblock


\bibitem[Dencker et~al\mbox{.}(2020)]%
        {dencker2020deep}
\bibfield{author}{\bibinfo{person}{Tobias Dencker}, \bibinfo{person}{Pablo
  Klinkisch}, \bibinfo{person}{Stefan~M Maul}, {and} \bibinfo{person}{Bj{\"o}rn
  Ommer}.} \bibinfo{year}{2020}\natexlab{}.
\newblock \showarticletitle{Deep learning of cuneiform sign detection with weak
  supervision using transliteration alignment}.
\newblock \bibinfo{journal}{\emph{Plos one}} \bibinfo{volume}{15},
  \bibinfo{number}{12} (\bibinfo{year}{2020}), \bibinfo{pages}{e0243039}.
\newblock


\bibitem[Fogel et~al\mbox{.}(2020)]%
        {fogel2020scrabblegan}
\bibfield{author}{\bibinfo{person}{Sharon Fogel}, \bibinfo{person}{Hadar
  Averbuch-Elor}, \bibinfo{person}{Sarel Cohen}, \bibinfo{person}{Shai Mazor},
  {and} \bibinfo{person}{Roee Litman}.} \bibinfo{year}{2020}\natexlab{}.
\newblock \showarticletitle{ScrabbleGAN: Semi-Supervised Varying Length
  Handwritten Text Generation}. In \bibinfo{booktitle}{\emph{CVPR}}.
  \bibinfo{pages}{4324--4333}.
\newblock


\bibitem[Gao et~al\mbox{.}(2019)]%
        {gao2019artistic}
\bibfield{author}{\bibinfo{person}{Yue Gao}, \bibinfo{person}{Yuan Guo},
  \bibinfo{person}{Zhouhui Lian}, \bibinfo{person}{Yingmin Tang}, {and}
  \bibinfo{person}{Jianguo Xiao}.} \bibinfo{year}{2019}\natexlab{}.
\newblock \showarticletitle{Artistic glyph image synthesis via one-stage
  few-shot learning}.
\newblock \bibinfo{journal}{\emph{ACM TOG}} \bibinfo{volume}{38},
  \bibinfo{number}{6} (\bibinfo{year}{2019}), \bibinfo{pages}{1--12}.
\newblock


\bibitem[Gao and Wu(2020)]%
        {gao2020gan}
\bibfield{author}{\bibinfo{person}{Yiming Gao} {and} \bibinfo{person}{Jiangqin
  Wu}.} \bibinfo{year}{2020}\natexlab{}.
\newblock \showarticletitle{GAN-Based Unpaired Chinese Character Image
  Translation via Skeleton Transformation and Stroke Rendering}. In
  \bibinfo{booktitle}{\emph{AAAI}}, Vol.~\bibinfo{volume}{34}.
  \bibinfo{pages}{646--653}.
\newblock


\bibitem[Gates(1978)]%
        {gates1978outline}
\bibfield{author}{\bibinfo{person}{William Gates}.}
  \bibinfo{year}{1978}\natexlab{}.
\newblock \bibinfo{booktitle}{\emph{An outline dictionary of Maya glyphs, with
  a concordance and analysis of their relationships: with the author's" Glyph
  studies" reprinted from the Maya Society quarterly}}.
\newblock \bibinfo{publisher}{Courier Corporation}.
\newblock


\bibitem[Gokaslan et~al\mbox{.}(2018)]%
        {improve_Shape_def}
\bibfield{author}{\bibinfo{person}{Aaron Gokaslan}, \bibinfo{person}{Vivek
  Ramanujan}, \bibinfo{person}{Daniel Ritchie}, \bibinfo{person}{Kwang~In Kim},
  {and} \bibinfo{person}{James Tompkin}.} \bibinfo{year}{2018}\natexlab{}.
\newblock \showarticletitle{Improving Shape Deformation in Unsupervised
  Image-to-Image Translation}. In \bibinfo{booktitle}{\emph{ECCV}}.
  \bibinfo{pages}{649--665}.
\newblock


\bibitem[Goodfellow et~al\mbox{.}(2014)]%
        {2014Generative}
\bibfield{author}{\bibinfo{person}{Ian~J. Goodfellow}, \bibinfo{person}{Jean
  Pouget-Abadie}, \bibinfo{person}{Mehdi Mirza}, \bibinfo{person}{Bing Xu},
  \bibinfo{person}{David Warde-Farley}, \bibinfo{person}{Sherjil Ozair},
  \bibinfo{person}{Aaron Courville}, {and} \bibinfo{person}{Yoshua Bengio}.}
  \bibinfo{year}{2014}\natexlab{}.
\newblock \showarticletitle{Generative Adversarial Nets}. In
  \bibinfo{booktitle}{\emph{NeurIPS}}, Vol.~\bibinfo{volume}{27}.
\newblock


\bibitem[Graves(2013)]%
        {graves2013generating}
\bibfield{author}{\bibinfo{person}{Alex Graves}.}
  \bibinfo{year}{2013}\natexlab{}.
\newblock \showarticletitle{Generating sequences with recurrent neural
  networks}.
\newblock \bibinfo{journal}{\emph{arXiv preprint arXiv:1308.0850}}
  (\bibinfo{year}{2013}).
\newblock


\bibitem[Guo et~al\mbox{.}(2019)]%
        {guo2019discriminative}
\bibfield{author}{\bibinfo{person}{Ting Guo}, \bibinfo{person}{Xingquan Zhu},
  \bibinfo{person}{Yang Wang}, {and} \bibinfo{person}{Fang Chen}.}
  \bibinfo{year}{2019}\natexlab{}.
\newblock \showarticletitle{Discriminative sample generation for deep
  imbalanced learning}. In \bibinfo{booktitle}{\emph{IJCAI}}.
\newblock


\bibitem[Hamdany et~al\mbox{.}(2021)]%
        {hamdany2021translating}
\bibfield{author}{\bibinfo{person}{Arwa Hamed~Salih Hamdany},
  \bibinfo{person}{Raid Rafi~Omar Al-Nima}, {and} \bibinfo{person}{Lubab~H
  Albak}.} \bibinfo{year}{2021}\natexlab{}.
\newblock \showarticletitle{Translating cuneiform symbols using artificial
  neural network}.
\newblock \bibinfo{journal}{\emph{Telkomnika}} \bibinfo{volume}{19},
  \bibinfo{number}{2} (\bibinfo{year}{2021}), \bibinfo{pages}{438--443}.
\newblock


\bibitem[Han et~al\mbox{.}(2021)]%
        {han2021dual}
\bibfield{author}{\bibinfo{person}{Junlin Han}, \bibinfo{person}{Mehrdad
  Shoeiby}, \bibinfo{person}{Lars Petersson}, {and}
  \bibinfo{person}{Mohammad~Ali Armin}.} \bibinfo{year}{2021}\natexlab{}.
\newblock \showarticletitle{Dual Contrastive Learning for Unsupervised
  Image-to-Image Translation}. In \bibinfo{booktitle}{\emph{CVPR}}.
  \bibinfo{pages}{746--755}.
\newblock


\bibitem[Heusel et~al\mbox{.}(2017)]%
        {heusel2017gans}
\bibfield{author}{\bibinfo{person}{Martin Heusel}, \bibinfo{person}{Hubert
  Ramsauer}, \bibinfo{person}{Thomas Unterthiner}, \bibinfo{person}{Bernhard
  Nessler}, {and} \bibinfo{person}{Sepp Hochreiter}.}
  \bibinfo{year}{2017}\natexlab{}.
\newblock \showarticletitle{Gans trained by a two time-scale update rule
  converge to a local nash equilibrium}. In
  \bibinfo{booktitle}{\emph{NeurIPS}}, Vol.~\bibinfo{volume}{30}.
\newblock


\bibitem[Huang et~al\mbox{.}(2019)]%
        {Shuangping2019OBC306}
\bibfield{author}{\bibinfo{person}{Shuangping Huang}, \bibinfo{person}{Haobin
  Wang}, \bibinfo{person}{Yongge Liu}, \bibinfo{person}{Xiaosong Shi}, {and}
  \bibinfo{person}{Lianwen Jin}.} \bibinfo{year}{2019}\natexlab{}.
\newblock \showarticletitle{OBC306: A Large-Scale Oracle Bone Character
  Recognition Dataset}. In \bibinfo{booktitle}{\emph{ICDAR}}.
  \bibinfo{pages}{681--688}.
\newblock


\bibitem[Huang et~al\mbox{.}(2020b)]%
        {chen2020reusing}
\bibfield{author}{\bibinfo{person}{Siyu Huang}, \bibinfo{person}{Haoyi Xiong},
  \bibinfo{person}{Zhi-Qi Cheng}, \bibinfo{person}{Qingzhong Wang},
  \bibinfo{person}{Xingran Zhou}, \bibinfo{person}{Bihan Wen},
  \bibinfo{person}{Jun Huan}, {and} \bibinfo{person}{Dejing Dou}.}
  \bibinfo{year}{2020}\natexlab{b}.
\newblock \showarticletitle{Generating Person Images with Appearance-aware Pose
  Stylizer}. In \bibinfo{booktitle}{\emph{IJCAI}}. \bibinfo{pages}{623--629}.
\newblock


\bibitem[Huang et~al\mbox{.}(2018)]%
        {huang2018multimodal}
\bibfield{author}{\bibinfo{person}{Xun Huang}, \bibinfo{person}{Ming-Yu Liu},
  \bibinfo{person}{Serge Belongie}, {and} \bibinfo{person}{Jan Kautz}.}
  \bibinfo{year}{2018}\natexlab{}.
\newblock \showarticletitle{Multimodal unsupervised image-to-image
  translation}. In \bibinfo{booktitle}{\emph{ECCV}}. \bibinfo{pages}{172--189}.
\newblock


\bibitem[Huang et~al\mbox{.}(2020a)]%
        {huang2020rd}
\bibfield{author}{\bibinfo{person}{Yaoxiong Huang}, \bibinfo{person}{Mengchao
  He}, \bibinfo{person}{Lianwen Jin}, {and} \bibinfo{person}{Yongpan Wang}.}
  \bibinfo{year}{2020}\natexlab{a}.
\newblock \showarticletitle{RD-GAN: Few/Zero-Shot Chinese Character Style
  Transfer via Radical Decomposition and Rendering}. In
  \bibinfo{booktitle}{\emph{ECCV}}. \bibinfo{pages}{156--172}.
\newblock


\bibitem[Isola et~al\mbox{.}(2017)]%
        {isola2017imagepix2}
\bibfield{author}{\bibinfo{person}{Phillip Isola}, \bibinfo{person}{Jun-Yan
  Zhu}, \bibinfo{person}{Tinghui Zhou}, {and} \bibinfo{person}{Alexei~A
  Efros}.} \bibinfo{year}{2017}\natexlab{}.
\newblock \showarticletitle{Image-to-image translation with conditional
  adversarial networks}. In \bibinfo{booktitle}{\emph{CVPR}}.
  \bibinfo{pages}{1125--1134}.
\newblock


\bibitem[Jaderberg et~al\mbox{.}(2015)]%
        {jaderberg2015spatial}
\bibfield{author}{\bibinfo{person}{Max Jaderberg}, \bibinfo{person}{Karen
  Simonyan}, \bibinfo{person}{Andrew Zisserman}, {et~al\mbox{.}}}
  \bibinfo{year}{2015}\natexlab{}.
\newblock \showarticletitle{Spatial transformer networks}. In
  \bibinfo{booktitle}{\emph{NeurIPS}}, Vol.~\bibinfo{volume}{28}.
  \bibinfo{pages}{2017--2025}.
\newblock


\bibitem[Ji and Chen(2019)]%
        {ji2019generative}
\bibfield{author}{\bibinfo{person}{Bo Ji} {and} \bibinfo{person}{Tianyi Chen}.}
  \bibinfo{year}{2019}\natexlab{}.
\newblock \showarticletitle{Generative adversarial network for handwritten
  text}.
\newblock \bibinfo{journal}{\emph{arXiv preprint arXiv:1907.11845}}
  (\bibinfo{year}{2019}).
\newblock


\bibitem[Jiang et~al\mbox{.}(2017)]%
        {jiang2017dcfont}
\bibfield{author}{\bibinfo{person}{Yue Jiang}, \bibinfo{person}{Zhouhui Lian},
  \bibinfo{person}{Yingmin Tang}, {and} \bibinfo{person}{Jianguo Xiao}.}
  \bibinfo{year}{2017}\natexlab{}.
\newblock \showarticletitle{DCFont: an end-to-end deep Chinese font generation
  system}.
\newblock In \bibinfo{booktitle}{\emph{SIGGRAPH Asia 2017 Technical Briefs}}.
  \bibinfo{pages}{1--4}.
\newblock


\bibitem[Jiang et~al\mbox{.}(2019)]%
        {jiang2019scfont}
\bibfield{author}{\bibinfo{person}{Yue Jiang}, \bibinfo{person}{Zhouhui Lian},
  \bibinfo{person}{Yingmin Tang}, {and} \bibinfo{person}{Jianguo Xiao}.}
  \bibinfo{year}{2019}\natexlab{}.
\newblock \showarticletitle{Scfont: Structure-guided chinese font generation
  via deep stacked networks}. In \bibinfo{booktitle}{\emph{Proceedings of the
  AAAI conference on artificial intelligence}}, Vol.~\bibinfo{volume}{33}.
  \bibinfo{pages}{4015--4022}.
\newblock


\bibitem[Kang et~al\mbox{.}(2020)]%
        {kang2020ganwriting}
\bibfield{author}{\bibinfo{person}{Lei Kang}, \bibinfo{person}{Pau Riba},
  \bibinfo{person}{Yaxing Wang}, \bibinfo{person}{Mar{\c{c}}al Rusi{\~n}ol},
  \bibinfo{person}{Alicia Forn{\'e}s}, {and} \bibinfo{person}{Mauricio
  Villegas}.} \bibinfo{year}{2020}\natexlab{}.
\newblock \showarticletitle{GANwriting: content-conditioned generation of
  styled handwritten word images}. In \bibinfo{booktitle}{\emph{European
  Conference on Computer Vision}}. Springer, \bibinfo{pages}{273--289}.
\newblock


\bibitem[Kingma and Ba(2014)]%
        {kingma2014adam}
\bibfield{author}{\bibinfo{person}{Diederik~P Kingma} {and}
  \bibinfo{person}{Jimmy Ba}.} \bibinfo{year}{2014}\natexlab{}.
\newblock \showarticletitle{Adam: A method for stochastic optimization}.
\newblock \bibinfo{journal}{\emph{arXiv preprint arXiv:1412.6980}}
  (\bibinfo{year}{2014}).
\newblock


\bibitem[Lee et~al\mbox{.}(2020)]%
        {lee2020drit++}
\bibfield{author}{\bibinfo{person}{Hsin-Ying Lee}, \bibinfo{person}{Hung-Yu
  Tseng}, \bibinfo{person}{Qi Mao}, \bibinfo{person}{Jia-Bin Huang},
  \bibinfo{person}{Yu-Ding Lu}, \bibinfo{person}{Maneesh Singh}, {and}
  \bibinfo{person}{Ming-Hsuan Yang}.} \bibinfo{year}{2020}\natexlab{}.
\newblock \showarticletitle{Drit++: Diverse image-to-image translation via
  disentangled representations}.
\newblock \bibinfo{journal}{\emph{IJCV}} \bibinfo{volume}{128},
  \bibinfo{number}{10} (\bibinfo{year}{2020}), \bibinfo{pages}{2402--2417}.
\newblock


\bibitem[Lin et~al\mbox{.}(2019)]%
        {park2019semantic}
\bibfield{author}{\bibinfo{person}{Jianxin Lin}, \bibinfo{person}{Yingce Xia},
  \bibinfo{person}{Yijun Wang}, \bibinfo{person}{Tao Qin}, {and}
  \bibinfo{person}{Zhibo Chen}.} \bibinfo{year}{2019}\natexlab{}.
\newblock \showarticletitle{Image-to-Image Translation with Multi-Path
  Consistency Regularization}. In \bibinfo{booktitle}{\emph{IJCAI}}.
  \bibinfo{pages}{2980--2986}.
\newblock


\bibitem[Lin and Wan(2007)]%
        {lin2007style}
\bibfield{author}{\bibinfo{person}{Zhouchen Lin} {and} \bibinfo{person}{Liang
  Wan}.} \bibinfo{year}{2007}\natexlab{}.
\newblock \showarticletitle{Style-preserving English handwriting synthesis}.
\newblock \bibinfo{journal}{\emph{PR}} \bibinfo{volume}{40},
  \bibinfo{number}{7} (\bibinfo{year}{2007}), \bibinfo{pages}{2097--2109}.
\newblock


\bibitem[Liu and Gao(2018)]%
        {soc_from}
\bibfield{author}{\bibinfo{person}{Guoying Liu} {and} \bibinfo{person}{Feng
  Gao}.} \bibinfo{year}{2018}\natexlab{}.
\newblock \showarticletitle{Oracle-Bone Inscription Recognition Based on Deep
  Convolutional Neural Network}.
\newblock \bibinfo{journal}{\emph{Journal of Computers}}  \bibinfo{volume}{13}
  (\bibinfo{year}{2018}), \bibinfo{pages}{1442--1450}.
\newblock


\bibitem[Liu et~al\mbox{.}(2021a)]%
        {liu2021improved}
\bibfield{author}{\bibinfo{person}{Zhiqiang Liu}, \bibinfo{person}{Chengkai
  Huang}, {and} \bibinfo{person}{Yanxia Liu}.}
  \bibinfo{year}{2021}\natexlab{a}.
\newblock \showarticletitle{Improved Knowledge Distillation via Adversarial
  Collaboration}.
\newblock \bibinfo{journal}{\emph{arXiv preprint arXiv:2111.14356}}
  (\bibinfo{year}{2021}).
\newblock


\bibitem[Liu et~al\mbox{.}(2021b)]%
        {liu2021deep}
\bibfield{author}{\bibinfo{person}{Zhuoman Liu}, \bibinfo{person}{Wei Jia},
  \bibinfo{person}{Ming Yang}, \bibinfo{person}{Peiyao Luo},
  \bibinfo{person}{Yong Guo}, {and} \bibinfo{person}{Mingkui Tan}.}
  \bibinfo{year}{2021}\natexlab{b}.
\newblock \showarticletitle{Deep View Synthesis via Self-Consistent Generative
  Network}.
\newblock \bibinfo{journal}{\emph{IEEE Transactions on Multimedia}}
  (\bibinfo{year}{2021}).
\newblock


\bibitem[Liu et~al\mbox{.}(2021c)]%
        {LiuLH21}
\bibfield{author}{\bibinfo{person}{Zhiqiang Liu}, \bibinfo{person}{Yanxia Liu},
  {and} \bibinfo{person}{Chengkai Huang}.} \bibinfo{year}{2021}\natexlab{c}.
\newblock \showarticletitle{Semi-Online Knowledge Distillation}. In
  \bibinfo{booktitle}{\emph{British Machine Vision Conference}}.
  \bibinfo{publisher}{{BMVA} Press}, \bibinfo{pages}{33}.
\newblock


\bibitem[Luo et~al\mbox{.}(2020)]%
        {luo2020learn}
\bibfield{author}{\bibinfo{person}{Canjie Luo}, \bibinfo{person}{Yuanzhi Zhu},
  \bibinfo{person}{Lianwen Jin}, {and} \bibinfo{person}{Yongpan Wang}.}
  \bibinfo{year}{2020}\natexlab{}.
\newblock \showarticletitle{Learn to augment: Joint data augmentation and
  network optimization for text recognition}. In
  \bibinfo{booktitle}{\emph{Proceedings of the IEEE/CVF Conference on Computer
  Vision and Pattern Recognition}}. \bibinfo{pages}{13746--13755}.
\newblock


\bibitem[Mao et~al\mbox{.}(2017)]%
        {LSGAN}
\bibfield{author}{\bibinfo{person}{Xudong Mao}, \bibinfo{person}{Qing Li},
  \bibinfo{person}{Haoran Xie}, \bibinfo{person}{Raymond~YK Lau},
  \bibinfo{person}{Zhen Wang}, {and} \bibinfo{person}{Stephen Paul~Smolley}.}
  \bibinfo{year}{2017}\natexlab{}.
\newblock \showarticletitle{Least squares generative adversarial networks}. In
  \bibinfo{booktitle}{\emph{ICCV}}. \bibinfo{pages}{2794--2802}.
\newblock


\bibitem[Panagopoulos et~al\mbox{.}(2008)]%
        {panagopoulos2008automatic}
\bibfield{author}{\bibinfo{person}{Michail Panagopoulos},
  \bibinfo{person}{Constantin Papaodysseus}, \bibinfo{person}{Panayiotis
  Rousopoulos}, \bibinfo{person}{Dimitra Dafi}, {and} \bibinfo{person}{Stephen
  Tracy}.} \bibinfo{year}{2008}\natexlab{}.
\newblock \showarticletitle{Automatic writer identification of ancient Greek
  inscriptions}.
\newblock \bibinfo{journal}{\emph{IEEE TPAMI}} \bibinfo{volume}{31},
  \bibinfo{number}{8} (\bibinfo{year}{2008}), \bibinfo{pages}{1404--1414}.
\newblock


\bibitem[Park et~al\mbox{.}(2020)]%
        {park2020contrastive}
\bibfield{author}{\bibinfo{person}{Taesung Park}, \bibinfo{person}{Alexei~A
  Efros}, \bibinfo{person}{Richard Zhang}, {and} \bibinfo{person}{Jun-Yan
  Zhu}.} \bibinfo{year}{2020}\natexlab{}.
\newblock \showarticletitle{Contrastive learning for unpaired image-to-image
  translation}. In \bibinfo{booktitle}{\emph{European Conference on Computer
  Vision}}. Springer, \bibinfo{pages}{319--345}.
\newblock


\bibitem[Richardson et~al\mbox{.}(2021)]%
        {richardson2021encoding}
\bibfield{author}{\bibinfo{person}{Elad Richardson}, \bibinfo{person}{Yuval
  Alaluf}, \bibinfo{person}{Or Patashnik}, \bibinfo{person}{Yotam Nitzan},
  \bibinfo{person}{Yaniv Azar}, \bibinfo{person}{Stav Shapiro}, {and}
  \bibinfo{person}{Daniel Cohen-Or}.} \bibinfo{year}{2021}\natexlab{}.
\newblock \showarticletitle{Encoding in style: a stylegan encoder for
  image-to-image translation}. In \bibinfo{booktitle}{\emph{CVPR}}.
  \bibinfo{pages}{2287--2296}.
\newblock


\bibitem[Richardson and Weiss(2018)]%
        {richardson2018gans}
\bibfield{author}{\bibinfo{person}{Eitan Richardson} {and}
  \bibinfo{person}{Yair Weiss}.} \bibinfo{year}{2018}\natexlab{}.
\newblock \showarticletitle{On gans and gmms}.
\newblock \bibinfo{journal}{\emph{arXiv preprint arXiv:1805.12462}}
  (\bibinfo{year}{2018}).
\newblock


\bibitem[Rusakov et~al\mbox{.}(2019)]%
        {rusakov2019generating}
\bibfield{author}{\bibinfo{person}{Eugen Rusakov}, \bibinfo{person}{Kai
  Brandenbusch}, \bibinfo{person}{Denis Fisseler}, \bibinfo{person}{Turna
  Somel}, \bibinfo{person}{Gernot~A Fink}, \bibinfo{person}{Frank Weichert},
  {and} \bibinfo{person}{Gerfrid~GW M{\"u}ller}.}
  \bibinfo{year}{2019}\natexlab{}.
\newblock \showarticletitle{Generating Cuneiform Signs with Cycle-Consistent
  Adversarial Networks}. In \bibinfo{booktitle}{\emph{Proceedings of the 5th
  International Workshop on Historical Document Imaging and Processing}}.
  \bibinfo{pages}{19--24}.
\newblock


\bibitem[Shi et~al\mbox{.}(2016)]%
        {shi2016robust}
\bibfield{author}{\bibinfo{person}{Baoguang Shi}, \bibinfo{person}{Xinggang
  Wang}, \bibinfo{person}{Pengyuan Lyu}, \bibinfo{person}{Cong Yao}, {and}
  \bibinfo{person}{Xiang Bai}.} \bibinfo{year}{2016}\natexlab{}.
\newblock \showarticletitle{Robust scene text recognition with automatic
  rectification}. In \bibinfo{booktitle}{\emph{CVPR}}.
  \bibinfo{pages}{4168--4176}.
\newblock


\bibitem[Sommerschield(2020)]%
        {sommerschield2020ralegh}
\bibfield{author}{\bibinfo{person}{Thea Sommerschield}.}
  \bibinfo{year}{2020}\natexlab{}.
\newblock \showarticletitle{Ralegh Radford Rome Awards: Restoring ancient text
  using machine learning: a case-study on Greek and Latin epigraphy}.
\newblock \bibinfo{journal}{\emph{Papers of the British School at Rome}}
  \bibinfo{volume}{88} (\bibinfo{year}{2020}), \bibinfo{pages}{387--388}.
\newblock


\bibitem[Suh et~al\mbox{.}(2022)]%
        {suh2022discriminative}
\bibfield{author}{\bibinfo{person}{Sungho Suh}, \bibinfo{person}{Paul
  Lukowicz}, {and} \bibinfo{person}{Yong~Oh Lee}.}
  \bibinfo{year}{2022}\natexlab{}.
\newblock \showarticletitle{Discriminative feature generation for
  classification of imbalanced data}.
\newblock \bibinfo{journal}{\emph{PR}}  \bibinfo{volume}{122}
  (\bibinfo{year}{2022}), \bibinfo{pages}{108302}.
\newblock


\bibitem[Sun et~al\mbox{.}(2017)]%
        {sun2017learning}
\bibfield{author}{\bibinfo{person}{Danyang Sun}, \bibinfo{person}{Tongzheng
  Ren}, \bibinfo{person}{Chongxun Li}, \bibinfo{person}{Hang Su}, {and}
  \bibinfo{person}{Jun Zhu}.} \bibinfo{year}{2017}\natexlab{}.
\newblock \showarticletitle{Learning to write stylized chinese characters by
  reading a handful of examples}.
\newblock \bibinfo{journal}{\emph{arXiv preprint arXiv:1712.06424}}
  (\bibinfo{year}{2017}).
\newblock


\bibitem[Szegedy et~al\mbox{.}(2017)]%
        {incepv42017}
\bibfield{author}{\bibinfo{person}{Christian Szegedy}, \bibinfo{person}{Sergey
  Ioffe}, \bibinfo{person}{Vincent Vanhoucke}, {and}
  \bibinfo{person}{Alexander~A. Alemi}.} \bibinfo{year}{2017}\natexlab{}.
\newblock \showarticletitle{Inception-v4, Inception-ResNet and the Impact of
  Residual Connections on Learning}. In \bibinfo{booktitle}{\emph{AAAI}}.
  \bibinfo{pages}{4278--4284}.
\newblock


\bibitem[Tian(2017)]%
        {zi2zi}
\bibfield{author}{\bibinfo{person}{Yuchen Tian}.}
  \bibinfo{year}{2017}\natexlab{}.
\newblock \showarticletitle{zi2zi: Master Chinese calligraphy with conditional
  adversarial networks}. In
  \bibinfo{booktitle}{\emph{https://kaonashi-tyc.github.io/2017/04/06/zi2zi.html}}.
\newblock


\bibitem[Wang et~al\mbox{.}(2002)]%
        {wang2002learning}
\bibfield{author}{\bibinfo{person}{Jue Wang}, \bibinfo{person}{Chenyu Wu},
  \bibinfo{person}{Ying-Qing Xu}, \bibinfo{person}{Heung-Yeung Shum}, {and}
  \bibinfo{person}{Liang Ji}.} \bibinfo{year}{2002}\natexlab{}.
\newblock \showarticletitle{Learning-based cursive handwriting synthesis}. In
  \bibinfo{booktitle}{\emph{Proceedings Eighth International Workshop on
  Frontiers in Handwriting Recognition}}. \bibinfo{pages}{157--162}.
\newblock


\bibitem[Wen et~al\mbox{.}(2021)]%
        {wen2021zigan}
\bibfield{author}{\bibinfo{person}{Qi Wen}, \bibinfo{person}{Shuang Li},
  \bibinfo{person}{Bingfeng Han}, {and} \bibinfo{person}{Yi Yuan}.}
  \bibinfo{year}{2021}\natexlab{}.
\newblock \showarticletitle{ZiGAN: Fine-grained Chinese Calligraphy Font
  Generation via a Few-shot Style Transfer Approach}. In
  \bibinfo{booktitle}{\emph{Proceedings of the 29th ACM International
  Conference on Multimedia}}. \bibinfo{pages}{621--629}.
\newblock


\bibitem[Wu et~al\mbox{.}(2020)]%
        {wu2020calligan}
\bibfield{author}{\bibinfo{person}{Shan-Jean Wu}, \bibinfo{person}{Chih-Yuan
  Yang}, {and} \bibinfo{person}{Jane Yung-jen Hsu}.}
  \bibinfo{year}{2020}\natexlab{}.
\newblock \showarticletitle{CalliGAN: Style and Structure-aware Chinese
  Calligraphy Character Generator}.
\newblock \bibinfo{journal}{\emph{arXiv preprint arXiv:2005.12500}}
  (\bibinfo{year}{2020}).
\newblock


\bibitem[Xie et~al\mbox{.}(2021)]%
        {xie2021dg}
\bibfield{author}{\bibinfo{person}{Yangchen Xie}, \bibinfo{person}{Xinyuan
  Chen}, \bibinfo{person}{Li Sun}, {and} \bibinfo{person}{Yue Lu}.}
  \bibinfo{year}{2021}\natexlab{}.
\newblock \showarticletitle{DG-Font: Deformable Generative Networks for
  Unsupervised Font Generation}. In \bibinfo{booktitle}{\emph{CVPR}}.
  \bibinfo{pages}{5130--5140}.
\newblock


\bibitem[Xu et~al\mbox{.}(2020)]%
        {xu2020cross}
\bibfield{author}{\bibinfo{person}{Haoming Xu}, \bibinfo{person}{Runhao Zeng},
  \bibinfo{person}{Qingyao Wu}, \bibinfo{person}{Mingkui Tan}, {and}
  \bibinfo{person}{Chuang Gan}.} \bibinfo{year}{2020}\natexlab{}.
\newblock \showarticletitle{Cross-modal relation-aware networks for
  audio-visual event localization}. In \bibinfo{booktitle}{\emph{Proceedings of
  the 28th ACM International Conference on Multimedia}}.
  \bibinfo{pages}{3893--3901}.
\newblock


\bibitem[Xu(2021)]%
        {Xuzhongshu}
\bibfield{author}{\bibinfo{person}{Zhongshu Xu}.}
  \bibinfo{year}{2021}\natexlab{}.
\newblock \bibinfo{booktitle}{\emph{Jia Gu Wen Zi Dian}}.
\newblock \bibinfo{publisher}{Sichuan Lexicographical Publishing House}.
\newblock


\bibitem[Yamauchi et~al\mbox{.}(2018)]%
        {Yamauchi:2018}
\bibfield{author}{\bibinfo{person}{Kenji Yamauchi}, \bibinfo{person}{Hajime
  Yamamoto}, {and} \bibinfo{person}{Wakaha Mori}.}
  \bibinfo{year}{2018}\natexlab{}.
\newblock \showarticletitle{Building A Handwritten Cuneiform Character
  Imageset}. In \bibinfo{booktitle}{\emph{LREC}}.
\newblock


\bibitem[Yang et~al\mbox{.}(2019)]%
        {yang2019tet}
\bibfield{author}{\bibinfo{person}{Shuai Yang}, \bibinfo{person}{Jiaying Liu},
  \bibinfo{person}{Wenjing Wang}, {and} \bibinfo{person}{Zongming Guo}.}
  \bibinfo{year}{2019}\natexlab{}.
\newblock \showarticletitle{Tet-gan: Text effects transfer via stylization and
  destylization}. In \bibinfo{booktitle}{\emph{Proceedings of the AAAI
  Conference on Artificial Intelligence}}, Vol.~\bibinfo{volume}{33}.
  \bibinfo{pages}{1238--1245}.
\newblock


\bibitem[Yu and Koltun(2015)]%
        {yu2015multi}
\bibfield{author}{\bibinfo{person}{Fisher Yu} {and} \bibinfo{person}{Vladlen
  Koltun}.} \bibinfo{year}{2015}\natexlab{}.
\newblock \showarticletitle{Multi-scale context aggregation by dilated
  convolutions}.
\newblock \bibinfo{journal}{\emph{arXiv preprint arXiv:1511.07122}}
  (\bibinfo{year}{2015}).
\newblock


\bibitem[Zeng et~al\mbox{.}(2021)]%
        {zeng2021strokegan}
\bibfield{author}{\bibinfo{person}{Jinshan Zeng}, \bibinfo{person}{Qi Chen},
  \bibinfo{person}{Yunxin Liu}, \bibinfo{person}{Mingwen Wang}, {and}
  \bibinfo{person}{Yuan Yao}.} \bibinfo{year}{2021}\natexlab{}.
\newblock \showarticletitle{Strokegan: Reducing mode collapse in chinese font
  generation via stroke encoding}. In \bibinfo{booktitle}{\emph{proceedings of
  AAAI}}, Vol.~\bibinfo{volume}{3}.
\newblock


\bibitem[Zhan et~al\mbox{.}(2019)]%
        {zhan2019spatial}
\bibfield{author}{\bibinfo{person}{Fangneng Zhan}, \bibinfo{person}{Hongyuan
  Zhu}, {and} \bibinfo{person}{Shijian Lu}.} \bibinfo{year}{2019}\natexlab{}.
\newblock \showarticletitle{Spatial fusion gan for image synthesis}. In
  \bibinfo{booktitle}{\emph{CVPR}}. \bibinfo{pages}{3653--3662}.
\newblock


\bibitem[Zhang et~al\mbox{.}(2020b)]%
        {zhang2020ai}
\bibfield{author}{\bibinfo{person}{Chongsheng Zhang}, \bibinfo{person}{Ruixing
  Zong}, \bibinfo{person}{Shuang Cao}, \bibinfo{person}{Yi Men}, {and}
  \bibinfo{person}{Bofeng Mo}.} \bibinfo{year}{2020}\natexlab{b}.
\newblock \showarticletitle{AI-Powered Oracle Bone Inscriptions Recognition and
  Fragments Rejoining.}. In \bibinfo{booktitle}{\emph{IJCAI}}.
  \bibinfo{pages}{5309--5311}.
\newblock


\bibitem[Zhang et~al\mbox{.}(2018a)]%
        {zhang2018unreasonable}
\bibfield{author}{\bibinfo{person}{Richard Zhang}, \bibinfo{person}{Phillip
  Isola}, \bibinfo{person}{Alexei~A Efros}, \bibinfo{person}{Eli Shechtman},
  {and} \bibinfo{person}{Oliver Wang}.} \bibinfo{year}{2018}\natexlab{a}.
\newblock \showarticletitle{The unreasonable effectiveness of deep features as
  a perceptual metric}. In \bibinfo{booktitle}{\emph{CVPR}}.
  \bibinfo{pages}{586--595}.
\newblock


\bibitem[Zhang et~al\mbox{.}(2019)]%
        {zhang2019whole}
\bibfield{author}{\bibinfo{person}{Yifan Zhang}, \bibinfo{person}{Hanbo Chen},
  \bibinfo{person}{Ying Wei}, \bibinfo{person}{Peilin Zhao},
  \bibinfo{person}{Jiezhang Cao}, \bibinfo{person}{Xinjuan Fan},
  \bibinfo{person}{Xiaoying Lou}, \bibinfo{person}{Hailing Liu},
  \bibinfo{person}{Jinlong Hou}, \bibinfo{person}{Xiao Han}, {et~al\mbox{.}}}
  \bibinfo{year}{2019}\natexlab{}.
\newblock \showarticletitle{From whole slide imaging to microscopy: Deep
  microscopy adaptation network for histopathology cancer image
  classification}. In \bibinfo{booktitle}{\emph{International Conference on
  Medical Image Computing and Computer-Assisted Intervention}}.
  \bibinfo{pages}{360--368}.
\newblock


\bibitem[Zhang et~al\mbox{.}(2020a)]%
        {zhang2020collaborative}
\bibfield{author}{\bibinfo{person}{Yifan Zhang}, \bibinfo{person}{Ying Wei},
  \bibinfo{person}{Qingyao Wu}, \bibinfo{person}{Peilin Zhao},
  \bibinfo{person}{Shuaicheng Niu}, \bibinfo{person}{Junzhou Huang}, {and}
  \bibinfo{person}{Mingkui Tan}.} \bibinfo{year}{2020}\natexlab{a}.
\newblock \showarticletitle{Collaborative unsupervised domain adaptation for
  medical image diagnosis}.
\newblock \bibinfo{journal}{\emph{IEEE Transactions on Image Processing}}
  \bibinfo{volume}{29} (\bibinfo{year}{2020}), \bibinfo{pages}{7834--7844}.
\newblock


\bibitem[Zhang et~al\mbox{.}(2018b)]%
        {zhang2018separating}
\bibfield{author}{\bibinfo{person}{Yexun Zhang}, \bibinfo{person}{Ya Zhang},
  {and} \bibinfo{person}{Wenbin Cai}.} \bibinfo{year}{2018}\natexlab{b}.
\newblock \showarticletitle{Separating style and content for generalized style
  transfer}. In \bibinfo{booktitle}{\emph{CVPR}}. \bibinfo{pages}{8447--8455}.
\newblock


\bibitem[Zhu et~al\mbox{.}(2017a)]%
        {zhu2017unpaired}
\bibfield{author}{\bibinfo{person}{Jun-Yan Zhu}, \bibinfo{person}{Taesung
  Park}, \bibinfo{person}{Phillip Isola}, {and} \bibinfo{person}{Alexei~A
  Efros}.} \bibinfo{year}{2017}\natexlab{a}.
\newblock \showarticletitle{Unpaired image-to-image translation using
  cycle-consistent adversarial networks}. In \bibinfo{booktitle}{\emph{ICCV}}.
  \bibinfo{pages}{2223--2232}.
\newblock


\bibitem[Zhu et~al\mbox{.}(2017b)]%
        {zhu2017toward}
\bibfield{author}{\bibinfo{person}{Jun-Yan Zhu}, \bibinfo{person}{Richard
  Zhang}, \bibinfo{person}{Deepak Pathak}, \bibinfo{person}{Trevor Darrell},
  \bibinfo{person}{Alexei~A Efros}, \bibinfo{person}{Oliver Wang}, {and}
  \bibinfo{person}{Eli Shechtman}.} \bibinfo{year}{2017}\natexlab{b}.
\newblock \showarticletitle{Multimodal Image-to-Image Translation by Enforcing
  Bi-Cycle Consistency}. In \bibinfo{booktitle}{\emph{NeurIPS}}.
  \bibinfo{pages}{465--476}.
\newblock


\end{thebibliography}

\clearpage
\title{Appendix}
\begin{appendix}
\section{Appendix}

\renewcommand{\thetable}{A\arabic{table}}
\renewcommand{\thefigure}{A\arabic{figure}}

\setcounter{table}{0}
\begin{table}[h]
    \caption{Architecture of GTG Components. `conv' denotes a convolutional layer, followed by the channel number. `convT' denotes a transposed convolutional layer, followed by the channel number. `BN' and `MP' denote a batch normalization layer and a max pooling layer, respectively. Adding the $N^2$ coordinates for TPS grid shifting and the 4  parameters for affine transformation, including rotation, scaling, and shifting, a total of $2N^2+4$ parameters are obtained, which need to be learned.}
\begin{tabular}{cc}\toprule
\textit{Encoder}  \& \textit{Predictor} &   $R_x$ \& $R_z$ \\ \midrule
input $1\times64\times64$        & input $1 \times (2N^2+4)$         \\      \midrule
conv-64,BN,ReLU,MP     & fc-($2N^2$+4),BN,ReLU       \\ \midrule
conv-128,BN,ReLU,MP  &  fc-1024,BN,ReLU  \\ \midrule
conv-64,BN,ReLU,MP   & fc-1024,BN,ReLU  \\ \midrule    
conv-16,BN,ReLU,MP    & fc-1024         \\ \midrule  
Reshape to $1\times1024$&    Reshape to $64\times4\times4$ \\ \midrule
fc-1024,BN,ReLU     &     convT-64,BN,ReLU,MP      \\ \midrule
fc-1024,BN,ReLU        &     convT-128,BN,ReLU,MP        \\ \midrule
fc-1024,BN,ReLU      &   convT-64,BN,ReLU,MP            \\ \midrule
fc-($\text2N^2$+4)            &   convT-1,BN,ReLU,MP              \\ \midrule
output $1\times(2N^2+4)$          &  output $1\times64\times64$   \\
\bottomrule
\end{tabular}
\label{fig:t1}
\end{table}

\setcounter{figure}{0}
\begin{figure}[h]
  \centering
  \includegraphics[width=1.0\linewidth]{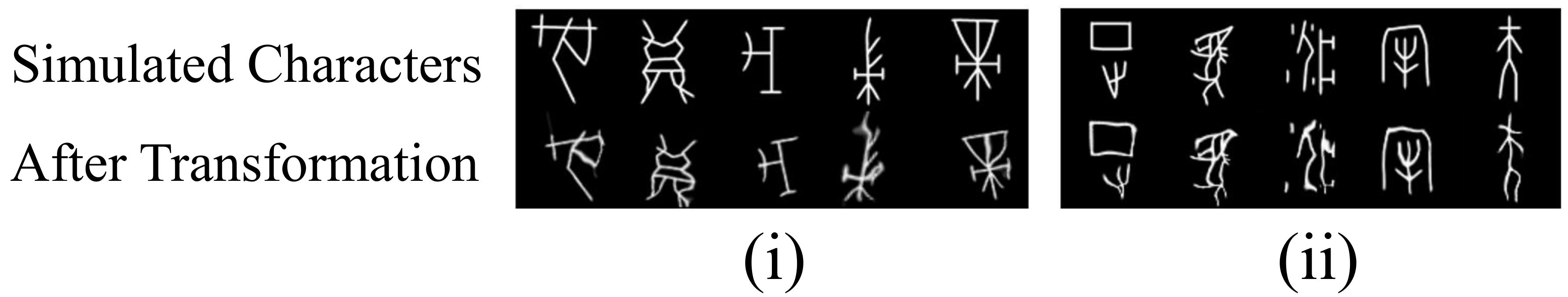}
   \caption{(i) Our TPS + affine transformation and (ii) only using TPS transformation.}
   \label{fig:transformation constrast}
\end{figure}

\begin{table}[h]
    \caption{Quantitative Evaluation of Generated PCCs.}
    \begin{center}
            \begin{tabular}{cccc} 
                \hline
                Methods  & NDB$\downarrow$  & JSD$\downarrow$ & LPIPS$\uparrow$ \\ 
                \hline
                DCLGAN~\cite{han2021dual}      & 16 & 0.342 & 0.220   \\
                DRIT++~\cite{lee2020drit++}        & 12 & 0.446 & 0.214\\
                NICE-GAN~\cite{chen2020reusing}      & 12 & 0.310 &  0.223\\
                CycleGAN~\cite{zhu2017unpaired}      & 11 & 0.174 & 0.231\\
                DG-Font~\cite{xie2021dg}     &  10  & 0.156 & 0.188       \\
                AGTGAN(Ours)       & {\bf 4} & {\bf 0.074}  & {\bf 0.253}\\ 
                \hline
            \end{tabular}
    \end{center}
    \label{tab:quantitative evaluation}
\end{table}

\subsection{Architecture Details}

The glyph shape generator $G_g$ contains a spatial transformer network (STN)~\cite{jaderberg2015spatial} component and two reconstruction networks, $R_x$ and $R_z$. Herein, the STN is used to predict the thin plate spline (TPS) grid coordinates and affine transformation parameters. $R_x$ and $R_z$ are used to restore the input glyph and the noise, respectively. The details of the \textit{Encoder}, \textit{Predictor} of STN, as well as $R_x$ \& $R_z$ are shown in Tab.~\ref{fig:t1}. The four convolutional layers in the first column constitute the \textit{Encoder}, and the other four fully connected layers, after the reshape operation, constitute the \textit{Predictor}. 
We build the signal reconstruction network $R_x$ using the inverse structure of the \textit{Encoder} \& \textit{Predictor}, while the structure of the noise reconstruction network $R_z$ is the same as the part of the fully connected layers in $R_x$.

\subsection{Effects of the combination of affine and TPS transformation}
Affine transformation can yield global deformations, e.g., rotation, translation, and scaling of characters, while TPS transformation tends to create local stroke modifications. It is demonstrated in Fig.~\ref{fig:transformation constrast} that the combination of affine and TPS transformations can yield global and local shape deformations simultaneously, while only using TPS transformation lacks  global shape deformation.

\subsection{Generated POCs of Unseen Character Classes}
\begin{figure*}
\begin{center}
    \includegraphics[width=1.0\linewidth]{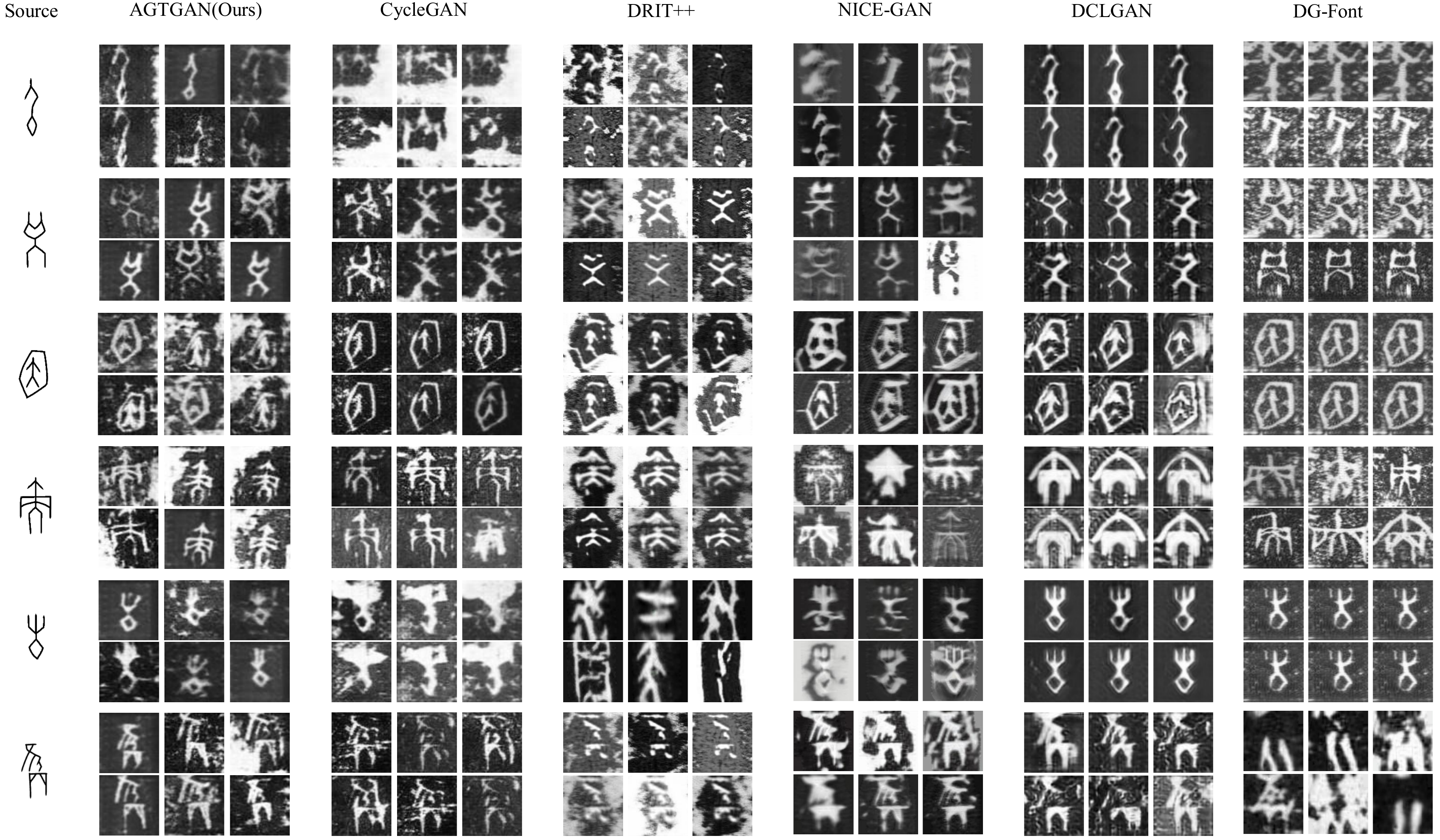}
\end{center}
   \caption{Generated POC images of unseen classes.}
\label{fig:f1}
\end{figure*}

Collecting paired ancient character data is costly or even impractical. Thanks to unsupervised learning, our method can generate samples that are unseen during training, which is of great significance to solve the problem of data scarcity.
Fig.~\ref{fig:f1} shows the comparison results between our AGTGAN and other models. Nine character classes unseen in OBC306 are randomly selected for comparison. We can see that our AGTGAN is still able to generate realistic and diverse photographic oracle bone characters (POCs), even if there are no real POC samples available for training. The POCs generated by other models are more or less flawed, as they do not have rich glyph shape variations, or have blurry strokes, fog effects, chaotic glyphs, etc.

\subsection{Quantitative Evaluation of Generated PCCs}
Tab.~\ref{tab:quantitative evaluation} summarizes the quantitative results of generated photographic cuneiform characters (PCCs). 
In terms of NDB and JSD, our method expresses the superior performance compared to other methods with a remarkable gap away from the second-best. Besides, our method also achieves the best results on LPIPS, which measures the diversity of generated samples.


\subsection{Classification Setting Details}
The classification performance is measured by the average class accuracy over all classes in the test set, rather than the accuracy of all samples. Otherwise, the overall accuracy will be dominated by the major classes. Using the the average class accuracy can equally reflect the accuracy of each character class, including the minority classes. For classes containing less than 750 training samples, we use the comparison methods to generate new POCs, to ensure that each class has at least 750 samples after combining the real and generated POCs. For classes with more than 750 samples, no extra generated images are used to train the classifier. It is worth noting that 750 is the average number of samples in the classes of OBC306. According to~\cite{Shuangping2019OBC306}, we select the best-performing Inception-v4 as the backbone of the POC recognizer.

\begin{table*}
    \caption{The class average accuracy achieved by different methods. `Source only' means training the recognizer without using any generated image.}
    \begin{center}
        \begin{tabular}{ccccccc} 
        \hline
                 & \multicolumn{3}{c}{TOP-3(\%)} & \multicolumn{3}{c}{TOP-5(\%)}  \\ 
                              \cline{2-7} 
        Methods  & OBC306  & FS  & ZS & OBC306  & FS  & ZS \\ 
                               \hline
        Source only   & 78.78       & 8.93    & 0  & 80.78       &  12.50   & 0 \\
        DG-Font           & 84.53      & 50.85          &37.25  & 87.42       & 58.40    & 47.06 \\
        CycleGAN         & 88.98      & 67.86        & 72.41  & 91.94       & 76.79    & 82.76 \\
        DRIT++           & 90.24     & 71.43        & 75.86  & 92.44       & 75.00    & 82.76 \\
        DCLGAN           & 91.31     & 77.47        & 74.51  &93.21       & 81.30    & 78.43  \\
        NICE-GAN          & 92.77       &82.14       &96.55  & 94.17       & 83.93    & 96.55 \\
        AGTGAN(Ours)     & {\bf 93.13}        & {\bf 85.71}   & {\bf 96.55}    & {\bf 94.61}   & {\bf 87.50}       & {\bf 96.55}   \\ 
                                \hline
       \end{tabular}
    \end{center}
    \label{tab:class average accuracy appendix}
\end{table*}

\subsection{Classification Performance in terms of TOP-K}
The TOP-3 and TOP-5 average class accuracy are tabulated in Tab.~\ref{tab:class average accuracy appendix}.
Training the classifier with generated samples added, we achieve the best TOP-3 and TOP-5 performances, with the average class accuracies reaching 93.13\% and 94.61\%, respectively.
These results are significantly better than those achieved by all other models and `Source Only', demonstrating the superiority of our method.
Besides, we can see that AGTGAN and NICE-GAN achieve the same TOP-3 and TOP-5 accuracy of 96.55\% under the setting of ZS. 
The reason is that there are only one test sample per class in 29 zero-shot classes of OBC306, and the prediction performance for 28 out of 29 classes, enhanced by these two methods, can reach 100\% in terms of both TOP-3 and TOP-5 classification accuracies. 
Therefore, the ZS accuracy can be calculated as follow:  
$$\frac{28 \times 100\%}{29} = 96.55\%.$$

\end{appendix}

\end{document}